\documentclass[10pt,journal,compsoc]{IEEEtran}
\ifCLASSOPTIONcompsoc
  \usepackage[nocompress]{cite}
\else
  \usepackage{cite}
\fi
\ifCLASSINFOpdf
\else
\fi
\usepackage{textcomp}
\usepackage{graphicx}
\usepackage[caption=false]{subfig}
\usepackage{caption}
\usepackage{amsmath}
\usepackage{amssymb}
\usepackage{dirtytalk}
\usepackage{bbm}
\usepackage{derivative}
\usepackage[pagebackref,breaklinks,colorlinks=true,linkcolor=red,citecolor=blue]{hyperref}

\usepackage{booktabs}
\usepackage{algorithm,algpseudocode}
\usepackage{mathtools}
\usepackage{enumitem}

\DeclarePairedDelimiterX{\norm}[1]{\lVert}{\rVert}{#1}
\usepackage[hang]{footmisc}
\usepackage{tabularx,ragged2e}
\usepackage{multirow}
\usepackage{booktabs}
\usepackage{nth}
\usepackage{graphicx}
\usepackage{chngcntr}
\usepackage{xcolor,colortbl}

\newcolumntype{C}{>{\Centering\arraybackslash}X}  
\newcolumntype{t}{>{\Centering\hsize=.6\hsize}X}
\newcolumntype{s}{>{\Centering\hsize=.5\hsize}X}
\newcolumntype{j}{>{\Centering\hsize=.16\hsize}X}
\newcolumntype{k}{>{\Centering\hsize=.3\hsize}X}
\newcolumntype{y}{>{\Centering\hsize=.12\hsize}X}
\newcolumntype{i}{>{\Centering\hsize=.08\hsize}X}
\newcolumntype{v}{>{\Centering\hsize=.04\hsize}X}
\algdef{SE}[SUBALG]{Indent}{EndIndent}{}{\algorithmicend\ }%
\algtext*{Indent}
\algtext*{EndIndent}

\usepackage{tikz}

\usepackage{amsthm}
\usepackage{amssymb}

\newtheorem{theorem}{Theorem}[section]

\begin{document}
\title{Efficient Task Grouping Through Sample-wise Optimisation Landscape Analysis}

\author{
Anshul Thakur$^{\ast}$, 
Yichen Huang$^{\ast}$,
Soheila Molaei,
Yujiang Wang$^{\dagger}$,  
and David A. Clifton
\IEEEcompsocitemizethanks{
\IEEEcompsocthanksitem{
Anshul Thakur and Soheila Molaei are with Department of Engineering Science, University of Oxford, UK (e-mail: {anshul.thakur@eng.ox.ac.uk}; {soheila.molaei@eng.ox.ac.uk}).}
\IEEEcompsocthanksitem{Yichen Huang is with Saint Catherine's College, University of Oxford, UK (e-mail: {yichen.huang@stcatz.ox.ac.uk}).}
\IEEEcompsocthanksitem{Yujiang Wang is with Oxford Suzhou Centre for Advanced Research, China (e-mail: {yujiang.wang@oscar.ox.ac.uk}).}
\IEEEcompsocthanksitem{David A. Clifton is with Department of Engineering Science, University of Oxford, UK, and also with Oxford Suzhou Centre for Advanced Research, China (e-mail: {david.clifton@eng.ox.ac.uk}).}
}
\thanks{$^{\ast}$Equal contributions}
\thanks{$^{\dagger}$Corresponding author}

}
\newcommand\copyrightnotice{%
\begin{tikzpicture}[remember picture,overlay]
\node[anchor=south,yshift=10pt] at (current page.south) {\fbox{\parbox{\dimexpr\textwidth-\fboxsep-\fboxrule\relax}{\textcopyright\ This work has been submitted to the IEEE for possible publication. Copyright may be transferred without notice, after which this version may no longer be accessible.
 }}};
\end{tikzpicture}%
}
\markboth{}%
{Efficient Task Grouping Through Sample-wise Optimisation Landscape Analysis}
\IEEEtitleabstractindextext{
\begin{abstract}


Shared training approaches, such as multi-task learning (MTL) and gradient-based meta-learning, are widely used in various machine learning applications, but they often suffer from negative transfer, leading to performance degradation in specific tasks. While several optimisation techniques have been developed to mitigate this issue for pre-selected task cohorts, identifying optimal task combinations for joint learning—known as task grouping—remains underexplored and computationally challenging due to the exponential growth in task combinations and the need for extensive training and evaluation cycles. This paper introduces an efficient task grouping framework designed to reduce these overwhelming computational demands of the existing methods. The proposed framework infers pairwise task similarities through a sample-wise optimisation landscape analysis, eliminating the need for the shared model training required to infer task similarities in existing methods. With task similarities acquired, a graph-based clustering algorithm is employed to pinpoint near-optimal task groups, providing an approximate yet efficient and effective solution to the originally NP-hard problem. Empirical assessments conducted on $8$ different datasets highlight the effectiveness of the proposed framework, revealing a five-fold speed enhancement compared to previous state-of-the-art methods. Moreover, the framework consistently demonstrates comparable performance, confirming its remarkable efficiency and effectiveness in task grouping.

\end{abstract}

}

\maketitle
\copyrightnotice
\IEEEdisplaynontitleabstractindextext
\IEEEpeerreviewmaketitle

\IEEEraisesectionheading{
\section{Introduction}
\label{sec:introduction}
}

\IEEEPARstart{M}{ulti-task} learning (MTL) has emerged as an influential paradigm in machine learning (ML), allowing for the optimisation of a unified model to handle multiple tasks concurrently \cite{vandenhende2021multi,zhang2021survey,Caruana1998}. In critical applications with low-latency requirements, ranging from real-time decision-making systems in autonomous vehicles \cite{ishihara2021multi} to responsive healthcare diagnostics \cite{hou2023mtdiag}, employing a multi-task model can significantly reduce latency compared to using separate models for each task \cite{guo2020learning}. Beyond latency reduction, the shared representation learning inherent in MTL can provide implicit regularisation and facilitate information transfer among tasks, thereby enhancing performance, especially in scenarios with limited training data \cite{Caruana1998,caruana1993multitask}.

\vspace{0.1cm}
Although MTL holds promise, it is not universally beneficial and, in specific scenarios, can lead to subpar performance and diminished data efficiency compared to learning tasks individually \cite{yu2020gradient}. This behaviour is referred as \emph{negative transfer}, and often signals the underlying optimisation issues such as conflicting task-specific gradients and large differences among gradient magnitudes of specific tasks \cite{liu2019loss}. Negative transfer implies that the optimisation landscapes of the tasks involved in MTL contradict each other, complicating joint training or learning an effective shared representation space among these tasks. Consequently, it becomes pivotal to avoid tasks that induce negative transfer within the joint objective to fully capitalise on the advantages offered by joint training \cite{standley2020tasks}. 

\vspace{0.1cm}
In addition to MTL, gradient-based meta-learning \cite{finn2017model,nichol2018reptile} presents another effective mechanism for jointly optimising a shared model to learn multiple tasks \cite{thakur2021dynamic}. Similar to MTL, the optimisation objective of meta-learning involves the weighted average of loss functions of multiple tasks \cite{wang2021bridging}. Therefore, frameworks based on meta-learning are also susceptible to gradient conflicts and negative transfer, as they also train a shared model aimed at providing an effective representation space for all tasks. Hence, the identification of tasks that result in negative transfer is equally essential for effective gradient-based meta-learning. However, identification of such tasks proves to be a non-trivial endeavour, hinging upon the intricate unravelling of latent structures learned by neural networks \cite{fifty2021efficiently}. 

\begin{figure*}[t]
    \centering
\includegraphics[scale=0.52,trim={0.5cm 0.5cm 0 0}]{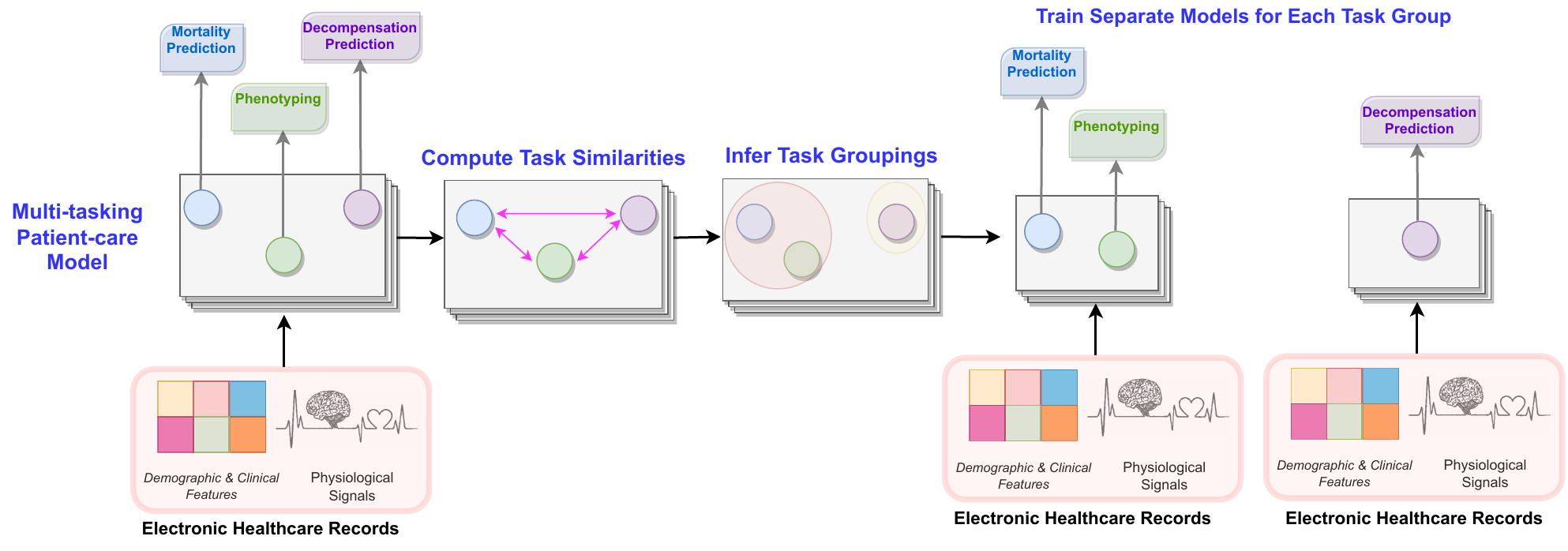}
    \caption{Illustration of a standard task grouping framework addressing three patient-care prediction tasks. A single multi-task model is replaced with two models where each model is trained for the identified task groupings. Contrary to this illustration, some task grouping frameworks doesn't allow singleton task groups.}
    \label{fig:framework}%
\end{figure*}

\vspace{0.1cm}
While most of the work targeting negative transfer deals with overcoming optimisation issues \cite{yu2020gradient,kendall2018multi,lee2016asymmetric}, a notable scarcity persists in studies focused on \emph{task groupings}—the pivotal identification of tasks that sidestep negative transfer and are capable of reinforcing each other through collaborative training \cite{kumar2012learning}. Figure \ref{fig:framework} illustrates a typical pipeline of a task grouping framework. Furthermore, the prominent task grouping approaches are computationally expensive and require training one or multiple multi-task models to establish relationships among tasks or determine which tasks should be grouped together for mutually beneficial joint optimisation \cite{standley2020tasks,fifty2021efficiently}. An initial approach to select ``optimum'' groupings from $\mathcal{T}$ tasks is based on exhaustive search over $2^{|\mathcal{T}|}-1$ multi-tasking models to compute pairwise task affinities, rendering the method almost impractical for a large number of tasks \cite{standley2020tasks}. Fifty \emph{et al.} \cite{fifty2021efficiently} introduced a substantial advancement over the exhaustive brute force approach by devising a method that entails training only a single multi-task model. However, this method still requires additional computationally intensive processing to evaluate the effect of each task-specific gradient update on the loss of other tasks during each training step, in order to infer task affinities or similarities. Identifying the optimal task groupings from these pairwise task affinities entails maximising the average affinities, a problem known to be NP-Hard \cite{standley2020tasks}. Both of these methods utilise branch-and-bound-like algorithms to address this challenge. While this approach yields satisfactory solutions within reasonable time frames for moderately sized task sets, its scalability becomes questionable when applied to larger sets of tasks. Therefore, there remains a need for scalable task grouping frameworks that demonstrate comparable performance to state-of-the-art approaches while demanding a fraction of the computation.

\vspace{0.1cm}
This paper introduces a task grouping framework that prioritises computational efficiency without compromising performance. It aims to rival state-of-the-art methods by delivering comparable performance while requiring only a fraction of the computational resources. Similar to state-of-the-art approaches, the proposed framework follows the pipeline outlined in Figure \ref{fig:framework}. It initiates by establishing task affinities and subsequently derives task groupings to train a multi-tasking model for each task group. However, in contrast to existing methods, the proposed framework does not necessitate training any multi-tasking model to quantify task affinities. Instead, it leverages \emph{sample-wise optimisation landscape analysis} \cite{zhang2021gradsign} to explore task interactions without performing any training. This sample-wise optimisation analysis enables the framework to estimate the impact of jointly optimising different tasks on the corresponding average training loss, thereby aiding in inferring task affinities (refer to Section \ref{ssec:sample_wise_mtl}).

\vspace{0.1cm}
Furthermore, instead of directly tackling the NP-Hard problem of maximising overall task affinities to obtain task groupings, the proposed framework re-frames it as a more efficient and still effective graph node clustering problem. In this framework, tasks are depicted as nodes within a graph, interconnected by weighted edges that signify task affinities. Graph attention networks (GATs) \cite{velivckovic2017graph} are leveraged to learn node representations by analysing higher-order interactions among nodes. This means that each node's representation is influenced not only by its immediate neighbours but also by nodes several hops away. As a result, these node representations encapsulate not just pairwise similarities, but rather holistic information about interactions among nodes. These representations are subsequently utilised to cluster nodes or tasks, where each cluster can be interpreted as a task grouping.

\vspace{0.1cm}
\noindent The major contribution of this paper are as follows:

\begin{itemize}
\itemsep 0.1em
    \item This paper presents a computationally efficient task grouping framework that exhibits either comparable or better performance that state-of-the-art approaches.

    \item A theoretical analysis is conducted to elucidate the connection between sample-wise optimisation landscape analysis and negative transfer in Multi-Task Learning (MTL).

    \item This paper also demonstrates the capability of the proposed framework to be easily utilised in identifying task groupings within gradient-based meta-learning frameworks.

    \item A comprehensive empirical evaluation encompassing $5$ distinct datasets is conducted to substantiate the computational efficiency and proficient performance of the proposed framework.
\end{itemize}

\vspace{0.1cm}
The rest of the paper is organised as: Section 2 discusses the prior art. Section 3 presents the problem formulation and sample-wise optimisation landscape analysis. Section 4 presents the proposed framework. Section 5 and 6 document the experimental setup and results, respectively. Section 7 concludes this paper.




\section{Earlier Studies}
\label{sec:back}
\noindent{\textsc{\textbf{Multi-task learning:}}} MTL is a vast realm and a large amount of work has been done to avoid negative transfer. The majority of approaches primarily focus on architectural enhancements and optimisation-based strategies. These architectural enhancements help in learning task-specific representations, rather than a rigid shared representation, that minimise the potential negative transfer. Cross-switch \cite{misra2016cross} and multi-gate mixture-of-experts MTL \cite{ma2018modeling} are prominent examples of such architectural improvements. On the other hand, optimisation-based strategies target negative transfer by strategies to adeptly learn weights of loss functions in joint objective and direct gradient manipulations. Kendall \emph{et al.}~\cite{kendall2018multi} computes adaptive task-specific loss weights based on the homoscedastic uncertainty of each task. Inspired by this work, various strategies are formulated to weigh losses by analysing varying task-specific learning speeds \cite{chen2018gradnorm,liu2019end,zheng2019pyramidal} as well as task performance improvement trends \cite{guo2018dynamic,jean2019adaptive}. In particular, Chen \emph{et al.}~\cite{chen2018gradnorm} proposed an algorithm to automatically adapt gradient norms to balance the task-specific losses in the joint objective.  Deviating from these loss weighting methods, some approaches directly process the gradients to alleviate gradient conflicts and hence, negative transfer. Yu \emph{et al.} \cite{yu2020gradient} proposed to alter a pair of the conflicting gradients by projecting each onto the normal plane of other and hence, alleviating the gradient interference. On the similar lines, Navon \emph{et al.} \cite{navon2022multi} models multi-task learning as a bargaining game (where each task is considered as a player) and exploits Nash equilibrium solution to weight each gradient during joint optimisation. This framework finds a joint optima that works for all tasks and does not allow any the training to be dominated by any particular task.

\vspace{0.1cm}
\noindent{\textsc{\textbf{Task groupings:}}} 
As discussed earlier, the identification of the optimum task groupings is a scarcely studied topic within the realm of joint optimisation of models. Standley \emph{et al.} \cite{standley2020tasks} is one of the earliest studies addressing task groupings in depth. Their method trains multi-task models for every possible task pairs and then, uses high-order approximations (HOA) to infer the performance for different possible task groupings. Finally, a branch-and-bound like algorithm is used to obtain $b$ task groupings ($b$ being the inference budget or allowed models) that maximise the overall performance. 
Although effective, this method is computationally expensive as it requires training multiple multi-task models and computationally expensive branch-and-bound algorithm to identify optimum task groups. Fifty \emph{et al.}\cite{fifty2021efficiently} proposed task affinity groupings (TAG) to overcome these high computational requirements. TAG requires training a single multi-tasking model and observes the impact of each task-specific gradient update on the loss of every other task to computes affinities between different task pairs. TAG also exploits a branch-and-bound like algorithm to identify the task groupings that maximise average inter-task affinity.  Song \emph{et al.}~\cite{song2022efficient} view task grouping as a meta-learning problem and devise a meta-learner, MTG-Net, to map task combinations into performance gains.

\vspace{0.1cm}
\noindent{\textsc{\textbf{Comparison with the proposed method:}}}  The proposed method differs fundamentally from previous studies in its approach to estimating task affinities and determining task groupings. Unlike earlier methods, our approach estimates task affinities through sample-wise optimisation analysis, eliminating the need for model training. Additionally, instead of relying on an approximate solution to an NP-Hard problem, we use GAT and a clustering mechanism to identify effective task groupings.

\section{Problem Definition and Background}
\subsection{Problem Definition}
Given a set of $T$ tasks $\mathcal{T}=\{\tau_1,\tau_2,\ldots, \tau_T\}$, we want to assign them to $k$ groups such that tasks in each group are well-suited for joint training. In other words, we want to construct a set of $k$ multi-task models, $\mathcal{M}=\{m_1,m_2,\ldots,m_k\}$ where each model corresponds to one distinct task group. We define inference budget $b \geq k$ as the number of models allowed to be used at the inference time for all $n$ tasks. There is no constraint for task groups to be mutually exclusive. Should a task be part of multiple groups, several models are trained for that specific task. The selection for deployment or inference hinges on the most successful model based on validation performance. The collective performance of task groupings can be characterised as $\sum_i^T \mathcal{P}(\tau_i|\mathcal{M})/T$, wherein $\mathcal{P}(\tau_i|\mathcal{M})$ is the accuracy of task $\tau_i$ using the set $\mathcal{M}$ of trained models.


\subsection{Sample-wise Optimisation Landscape Analysis}
\label{sec:sample_wise_review}

\noindent \textsc{\textbf{Sample-wise local optima:}} 
Let $f_{\theta}(\cdot)$ be a neural network parameterised by $\theta \in \mathbb{R}^m$, and $\theta_0$ is the initial random state of the model. Then, for $i$-th training sample $(\mathbf{x}_i,y_i)$ in dataset $\mathcal{D}$, a sample-wise local optima $\theta^{\star}_i$ can be obtained using gradient descent:
\begin{equation}
    \theta^{\star}_i=\theta_0-\nabla_{\theta} \ell(f_{\theta}(\mathbf{x}_i),y_i), 
    \label{eq:sample}
\end{equation}

where $\ell(\cdot)$ is a loss function.

Since modelling capacity of neural networks is generally very high, it can be arguably assumed that one gradient update is sufficient to learn a single sample $(\mathbf{x}_i, y_i)$ i.e.~$\ell(f_{\theta^{\star}_{i}}(\mathbf{x}_i),y_i) \approx 0$ or to obtain sample-wise local optima $\theta_i^{\star}$ from an initial state $\theta_0$ \cite{zhang2021gradsign,yang2022towards}. As a result, at local optimum $\theta^{\star}_i$ corresponding to $\mathbf{x}_i$, we have $\nabla_{\theta} \ell(f_{\theta}(\mathbf{x}_i),y_i)|_{\theta^{\star}_i}=0$ and hence, local optimum is also global as it is impossible to achieve any lower loss for $\mathbf{x}_i$. Moreover, there are no saddle points in a sample-wise optimisation landscape as $\nabla^2_{\theta} \ell(f_{\theta}(\mathbf{x}_i),y_i)|_{\theta_i^{\star}}$ is always positive semi-definite \cite{zhang2021gradsign}.

\vspace{0.2cm}
\noindent \textsc{\textbf{Optimisation landscape analysis:}} The traditional approach to optimisation landscape analysis focuses on evaluating an objective across a mini-batch of training samples. In contrast, the sample-wise optimisation analysis breaks down the mini-batch objective into a collection of individual sample-wise optimisation landscapes. This approach has the potential to reveal hidden properties of optimisation landscapes that might be overlooked by traditional mini-batch analysis. Zhang and Jia \cite{zhang2021gradsign} highlighted that under reasonable assumptions, the density of sample-wise local optima, $\{\theta^{\star}_i\}_{i=1}^n$, corresponding to individual samples $\{(\mathbf{x}_i,y_i)\}_{i=1}^n$ can provide bounds on both the training and generalisation error of a neural network. In simpler terms, a higher density of sample-wise local optima indicates lower training and generalisation errors, and vice versa. Furthermore, Yang \emph{et al.} \cite{yang2022towards} leveraged sample-wise optimisation analysis to learn neural network initialisation that leads to improved convergence and generalisation. Thus, the sample-wise optimisation analysis presents a novel perspective on understanding optimisation landscapes in deep learning, providing insights that can lead to improved training, generalisation, and convergence of neural networks. \\

\section{Proposed Method}
This section presents the sample-wise optimisation landscape analysis in MTL setting. Subsequently, leveraging the insights gleaned from this sample-wise optimisation analysis, we present the proposed task grouping framework.

\subsection{Sample-wise optimisation landscape in multi-tasking}
\label{ssec:sample_wise_mtl}

A MTL dataset consisting of $n$ samples and $T$ learning tasks can be denoted as $\mathcal{D}=\{\mathbf{x}_i, (y_i^1,y_i^2,\ldots, y_i^T)\}_{i\in[n]}$ where $\mathbf{x}_i$ denotes the $i$-th sample and $y_i^t$ refers to its label corresponding to the $t$-th task.  In this work, we adopt the commonly used hard parameter-sharing multi-tasking neural architecture. We denote such a network as $f_{{\theta,\phi_t}}(\cdot)$, where it is parameterised by shared parameters $\theta \in \mathbb{R}^M$ and task-specific parameters $\phi_t \in \mathbb{R}^P$ corresponding to the $t$-th task. For the $i$-th sample $\mathbf{x}_i$, the loss from the $t$-th task can be written as $\mathcal{J}_i^t = \ell_t(f_{\{\theta,\phi_t\}}(\mathbf{x}_i),y_i^t)$, and the average sample-wise loss across all tasks is computed as:

\begin{align}
    \mathcal{J}_i &= \frac{1}{T} \sum_{t=1}^T \mathcal{J}_i^t = \frac{1}{T} \sum_{t=1}^T \ell_t(f_{\{\theta,\phi_t\}}(\mathbf{x}_i),y_i^t).
    \label{eq:loss}
\end{align}%

Joint optimisation in MTL or meta-learning primarily focuses on the shared parameters $\theta$, with the optimisation landscape of $\theta$ being crucial for studying task interactions. While the training of $\theta$ depends on task-specific parameters $\phi_t$, optimising $\phi_t$ themselves is straightforward. Therefore, this study primarily concentrates on analysing the loss landscape of $\mathbf{\theta}$. For brevity, we denote the multi-task network as $f_{\theta}(\cdot)$ instead of $f_{{\theta,\phi_t}}(\cdot)$ when appropriate.

\vspace{0.1cm}
For establishing an association between sample-wise optimisation landscape analysis and joint training, we decompose the process of learning from the $i$-th MTL sample $\left(\mathbf{x}_i,\{y_i^t\}_{t=1}^T\right)$ into learning from $T$ different task-specific samples $\{(\mathbf{x}_i,y_i^t)\}_{t=1}^T$. The initial random state of shared parameters $\theta_0$ is updated independently by each of these $T$ samples to obtain $T$ \emph{task-specific sample-wise} optima, i.e., $\{\theta_i^{t\star}\}_{t=1}^T$ (using Equation \ref{eq:sample}). Then, the shared optimum $\theta^{\star}_i$ for $i$-th MTL sample $\left(\mathbf{x}_i,\{y_i^t\}_{t=1}^T\right)$ across all tasks is the average of task-specific sample-wise optima $\{\theta_i^{t\star}\}_{t=1}^T$, as described in Theorem \ref{th:1}.

\vspace{0.2cm}
\begin{theorem}
Let $\left(\mathbf{x}_i,\{y_i^t\}_{t=1}^T\right)$ be a sample for updating the shared parameters, initialised with $\mathbf{\theta}_0$, in multi-tasking setup. Then, sample-wise gradient to update $\mathbf{\theta}_0$ is average of the task-specific sample-wise gradients i.e.~$\frac{1}{T} \sum_{t=1}^T \nabla_{\theta} \ell_t$. By extension, the shared optimum $\theta_i^{\star}$ is an average of the task-specific optima $\theta_i^{t\star}$, namely $\theta_i^{\star} = \frac{1}{T} \sum_{t=1}^T \theta_i^{t\star}$.
\label{th:1}
\end{theorem}
\renewcommand\qedsymbol{$\blacksquare$}
\begin{proof}

For simplicity, let $p_i^t$ denote predictions obtained for $\mathbf{x}_i$ for the $t$-th task, i.e., $p_i^t = f_{\{\theta, \phi_t\}}(\mathbf{x}_i)$. Following Equation \ref{eq:loss}, the average sample loss across all tasks can be written as $\mathcal{J}_i= \frac{1}{T} \sum_{t=1}^T \ell_t(p_i^t,y_i^t)$. Therefore, the sample-wise gradients w.r.t. $\mathbf{x}_i$ for updating $\mathbf{\theta}$ can be described as
\begin{equation}
    \nabla_{\theta}\mathcal{J}_i= \nabla_{\theta}\frac{1}{T} \sum_{t=1}^T \ell_t(p_i^t,y_i^t)= \frac{1}{T} \sum_{t=1}^T \nabla_{\theta} \ell_t(p_i^t,y_i^t).
\end{equation} 
The shared optimum $\theta_i^{\star}$ can be computed as
\begin{align}
    \theta_i^{\star} &=\theta_0 - \eta \nabla_{\theta}\mathcal{J}_i, \nonumber
    \\ &= \theta_0 - \eta \frac{1}{T} \sum_{t=1}^T \nabla_{\theta} \ell_t(p_i^t,y_i^t), \nonumber
    \\ &=\frac{1}{T} \sum_{t=1}^T(\theta_0 - \eta  \nabla_{\theta} \ell_t(p_i^t,y_i^t))
    \label{eq:shared_optimum}
\end{align}
where $\eta$ denotes the learning rate.

\vspace{0.1cm}
\noindent Notably, for the $t$-th task, the task-specific optima $\theta_i^{t\star}$ can be obtained as 
\begin{align}
\theta_i^{t\star} &= \theta_0 - \eta \nabla_{\theta}\mathcal{J}_i^t, \nonumber \\
&= \theta_0 - \eta  \nabla_{\theta} \ell_t(p_i^t,y_i^t). \label{eq:task_spec_optima}
\end{align}
Substituting Equation \ref{eq:task_spec_optima} into Equation \ref{eq:shared_optimum}, we have 
\begin{equation}
\theta_i^{\star} = \frac{1}{T} \sum_{t=1}^T \theta_i^{t\star}.  \label{eq:optimum}
\end{equation}

\end{proof}

\begin{figure}[t]
    \centering
\includegraphics[scale=0.48,trim={2cm 1cm 0 10cm}]{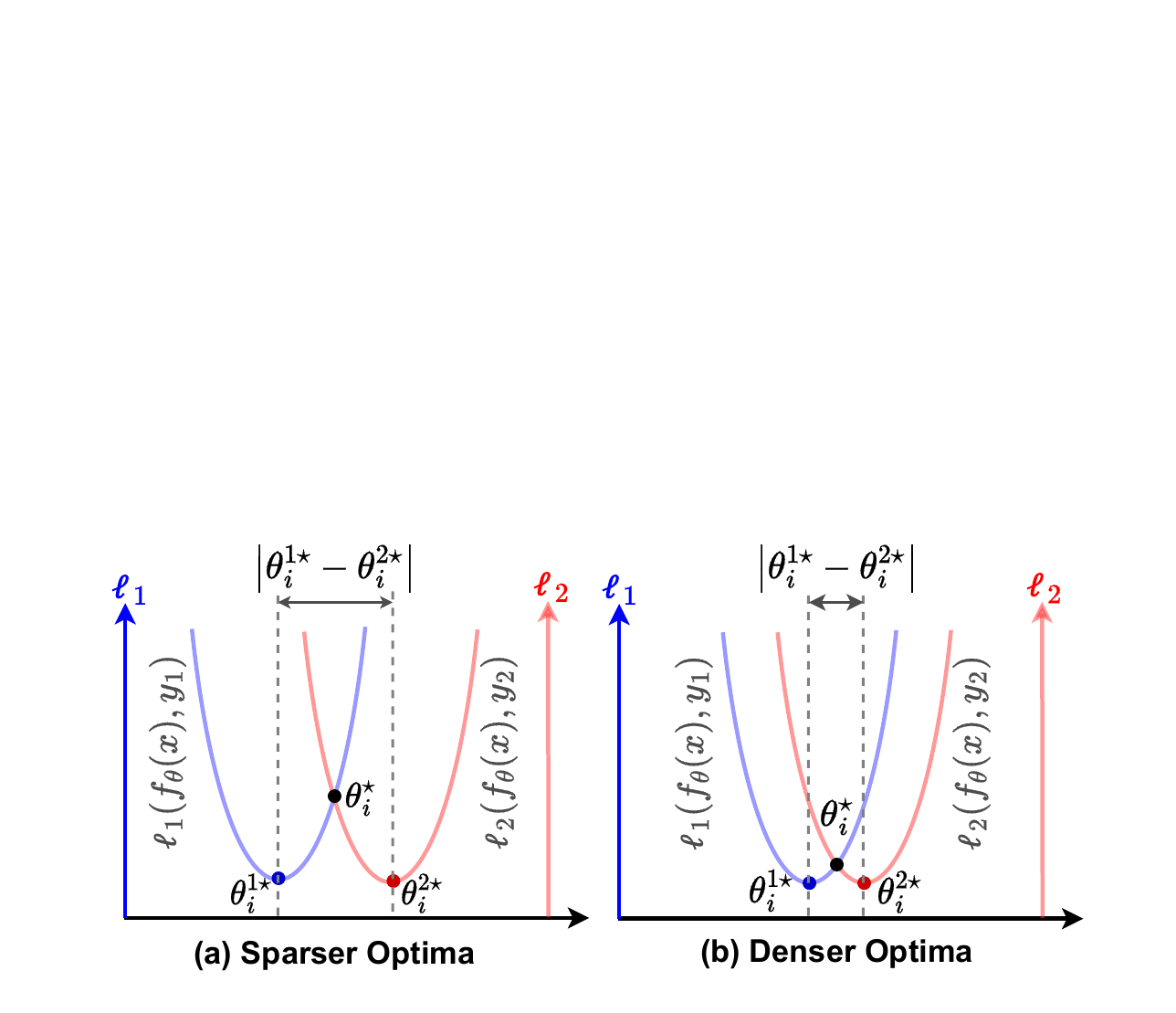} %
\caption{Illustration of the nature of shared global optima $\theta^{\star}$ as a function of the density of task-specific sample-wise local optima, $\theta_1^{\star}$ and $\theta_2^{\star}$ for tasks 1 and 2, respectively. In comparison to sparser local optima \textbf{(a)}, denser local optima \textbf{(b)} results in $\theta^{\star}$ that leads to better average loss across both tasks, $\mathcal{J}=(\ell_1+\ell_2)/2$, as desired in multi-tasking and other information sharing frameworks.}
\label{fig:sample_wise}
\end{figure}

Theorem \ref{th:1} implies that the shared optimum $\theta_i^{\star}$ is an average, or formally, a convex combination with equal coefficients of task-specific local optima $\{\theta_i^{t\star}\}_{t=1}^T$. The density of these task-specific optima, denoted as $\psi_{{\theta}_i}$, indicates the proximity of these task-specific optima among themselves and also describes their closeness to $\theta_i^{\star}$. This density $\psi_{{\theta}_i}$ is essentially the average distance between each pair of the task-specific sample-wise optima, and it can be formally written as 
\begin{equation}
       \psi_{{\theta}_i}=\frac{\sqrt{\mathcal{H}}}{T^2} \sum_{j=1}^{T}  \sum_{k=1}^{T}  \lVert \theta_i^{j\star}-\theta_i^{k\star}\rVert_1  
 \label{eq:dens}   
\end{equation}
where $\mathcal{H} \in \mathbb{R}$ is smoothness upper bound: $\forall k \in [M], t \in [T], [\nabla_{\theta}^2 \ell_t(p^t,y^t)]_{k,k} \leq \mathcal{H}$.

\vspace{0.1cm}
In the context of MTL, a higher density (and thus, a smaller $\psi_{{\theta}_i}$) implies that task-specific sample-wise optima $\{\theta_i^{t\star}\}_{t=1}^T$ are generally closer to each other; consequently, the shared optimum $\theta_i^{\star}$ is also closer to all task-specific optima, leading to a smaller and more desirable average loss.
Figure \ref{fig:sample_wise} provides an intuitive example to understand these observations, and we formally describe them in Theorem \ref{th:2}.


\vspace{0.15cm}
\begin{theorem}
For a set of $T$ tasks, the average training loss $\mathcal{J}_i$ for a sample $\left(\mathbf{x}_i,\{y_i^t\}_{t=1}^T\right)$ is upper bounded by the density $\psi_{\theta_i}$ of the task-specific sample-wise optima $\{\theta_i^{t\star}\}_{t=1}^T$, expressed as $\mathcal{J}_i\leq T^3\psi_{\theta_i}^2$.
\label{th:2}
\end{theorem}
\renewcommand\qedsymbol{$\blacksquare$}
\begin{proof}
$\theta_i^{\star}$ is the shared optimum obtained for $i$-th sample by aggregating task-specific sample-wise optima $\{\theta_i^{t\star}\}_{t=1}^T$ as defined in Equation \ref{eq:optimum}. Using standard results about smoothness, we obtain the following inequality 
\begin{align}
    \ell_t(f_{\theta_i^{\star}}(\mathbf{x}_i),y_i^t) \leq  &\ell_t(f_{\theta_i^{t\star}}(\mathbf{x}_i),y_i^t) \nonumber \\ &+ \nabla_{\theta} \ell_t(f_{\theta_i^{t\star}}(\mathbf{x}_i),y_i^t)^{\top}(\theta_i^{\star}-\theta_i^{t\star}) \nonumber \\ &+\frac{1}{2}\nabla_{\theta}^2 \ell_t(f_{\theta_i^{t\star}}(\mathbf{x}_i),y_i^t)^{\top}(\theta_i^{\star}-\theta_i^{t\star})^2.
    \label{eq:start}
\end{align}

Recalling that $\mathcal{H}$ is the smoothness upper bound satisfying $\nabla_{\theta}^2 \ell_t(f_{\theta}(\mathbf{x}_i),y_i^t)|_{\theta_i^{t\star}} \preceq \mathcal{H}\mathbb{I}$, and $\theta_i^{t\star}$ is a task-specific optimum with zero loss and also zero first-order derivatives, i.e., $\ell_t(f_{\theta_i^{t\star}}(\mathbf{x}_i),y_i^t)=0$ and $\nabla_{\theta} \ell_t(f_{\theta}(\mathbf{x}_i),y_i^t)|_{\theta_i^{t\star}}=0$. Then, Equation \ref{eq:start} can be simplified as
\begin{equation}
      \ell_t(f_{\theta_i^{\star}}(\mathbf{x}_i),y_i^t) \leq  \frac{1}{2}\mathcal{H} \lVert \theta_i^{\star}-\theta_i^{t\star} \rVert^2_2.
\end{equation} 

\noindent Averaging over all $T$ tasks on both sides, we have
\begin{align}
    \mathcal{J}_i = \frac{1}{T} \sum_{t} \ell_t(f_{\theta_i^{\star}}(\mathbf{x}_i),y_i^t) & \leq \frac{1}{2T} \sum_{t=1}^T\mathcal{H} \lVert \theta_i^{\star}-\theta_i^{t\star} \rVert^2_2 \nonumber \\
       &\leq \frac{\mathcal{H}}{T} \sum_{t=1}^T \lVert \theta_i^{\star} - \theta_i^{t\star} \rVert^2_1. \label{eq:inequality_expand}
\end{align}

\noindent Since $\mathbf{\theta}_i^{\star}$ is convex combination of $\{\theta_i^{t\star}\}_{t=1}^T$, namely $\mathbf{\theta}_i^{\star} = \sum_{j=1}^{T} \theta_i^{j\star}$, Equation \ref{eq:inequality_expand} can be written as

\begin{align}
        \mathcal{J}_i  \leq \frac{\mathcal{H}}{T} \sum_{t=1}^{T} \lVert \sum_{j=1}^{T} \theta_i^{j\star} - \theta_i^{t\star} \rVert^2_1
         & \leq \frac{\mathcal{H}}{T} \sum_{t=1}^{T} \sum_{j=1}^{T} \lVert \theta_i^{j\star} - \theta_i^{t\star} \rVert^2_1  \nonumber \\
    & \leq  T^3 \psi_{\theta_i}^2.
    \label{eq:end}
\end{align}
\end{proof}

\vspace{0.2cm}
Theorem \ref{th:2} highlights that a higher density, represented by a smaller $\psi_{\theta_i}$, results in a lower upper bound on the average training loss for $i$-th sample in MTL setup. By extension, the average multi-task loss across all $n$ samples in $\mathcal{D}$ can be described as
\begin{align}
  \mathcal{J}= \frac{1}{n} \sum_{i=1}^n \mathcal{J}_i & \leq  T^3\frac{1}{n} \sum_{i=1}^n \psi_{\theta_i}^2.
  \label{eq:left}
\end{align}

If we go one level deeper, let $\theta^{\star}=\frac{1}{n}\sum_{i=1}^n\theta_i^{\star}$ be the average shared optimum across $\mathcal{D}$ (all $T$ tasks and all $n$ training samples). Following Equation \ref{eq:dens}, the density among sample-wise optima $\{\theta_i^{\star}\}_{i=1}^n$ can be computed as

\begin{equation}
    \Psi_{{\theta^{\star}}}=\frac{\sqrt{\mathcal{H}}}{n^2} \sum_{j=1}^n \sum_{k=1}^n \lVert \theta_j^{\star}-\theta_k^{\star}\rVert_1.
    \label{eq:density_optimum}
\end{equation}

Following similar proofs in Equations \ref{eq:start}-\ref{eq:end}, we arrive at 
\begin{align}
  \mathcal{J}= \frac{1}{n} \sum_{i=1}^n \mathcal{J}_i & \leq  n^3 \Psi_{\theta^{\star}}^2.
  \label{eq:right}
\end{align}

Adding the inequalities in Equations \ref{eq:left} and \ref{eq:right} leads to 
\begin{align}
     \mathcal{J} & \leq \frac{T^3}{2n} \sum_{i=1}^n \psi_{\theta_i}^2 +  \frac{n^3}{2} \Psi_{\theta^{\star}}^2.
     \label{eq:last}
\end{align}

Equation \ref{eq:last} implies that MTL training loss $\mathcal{J}$ for $\mathcal{D}$ is upper-bounded by an aggregate of the average density of task-specific optima across all samples ($\frac{1}{n} \sum_{i=1}^n \psi_{\theta_i}^2$) and the density of sample-wise optima ($\Psi_{\theta^{\star}}^2$); higher densities ($\frac{1}{n} \sum_{i=1}^n \psi_{\theta_i}^2$ and $\Psi_{\theta^{\star}}^2$) implies better or lesser training loss.

\vspace{0.25cm}
\noindent \textsc{\textbf{Extension to meta learning:}}
The optimisation landscapes of multi-task learning and gradient-based meta-learning share similarities \cite{wang2021bridging}, and Theorem \ref{th:1} holds for both, indicating that shared optima across tasks are an aggregation of task-specific optima. In multi-task learning, a training unit comprises a single sample with multiple task labels $(\mathbf{x}_i,\{y_i^t\}_{t=1}^T)$, while in gradient-based meta-learning such as Reptile \cite{nichol2018reptile} with a single inner iteration, the $i$-th training unit can be viewed as $\{(\mathbf{x}_i^1,y_i^1),(\mathbf{x}_i^2,y_i^2),\ldots,(\mathbf{x}_i^T,y_i^T)\}$, with $(\mathbf{x}_i^t,y_i^t)$ as the training sample for the $t$-th task. Task-specific optima $\{\theta_i^{t\star}\}_{t=1}^T$ are obtained for each $t$-th task in the $i$-th meta-learning unit. Following Theorem \ref{th:2}, $\psi_{{\theta}_i}$, the density of$\{\theta_i^{t\star}\}_{t=1}^T$ , provides an upper bound on the average training loss for the $i$-th training unit in gradient-based meta-learning. Similarly, the average training loss for the entire dataset $\mathcal{D}$ is upper-bounded by an aggregate of $\Psi_{\theta^\star}$ and $\sum_{i=1}^n\psi_{{\theta}_i}$ (Equation \ref{eq:last}).



\subsection{Proposed Task Grouping Framework}
As illustrated in Figure \ref{fig:framework}, the proposed framework consists of two stages: inferring inter-task similarities and grouping or clustering tasks based on the inferred similarities. These stages are discussed below.

\begin{figure}[t]
    \centering
    \centering
    \subfloat[t-SNE representation of sample-wise converged models\centering]{{\includegraphics[scale=0.27,trim={0.5cm 0.5cm 0 0}]{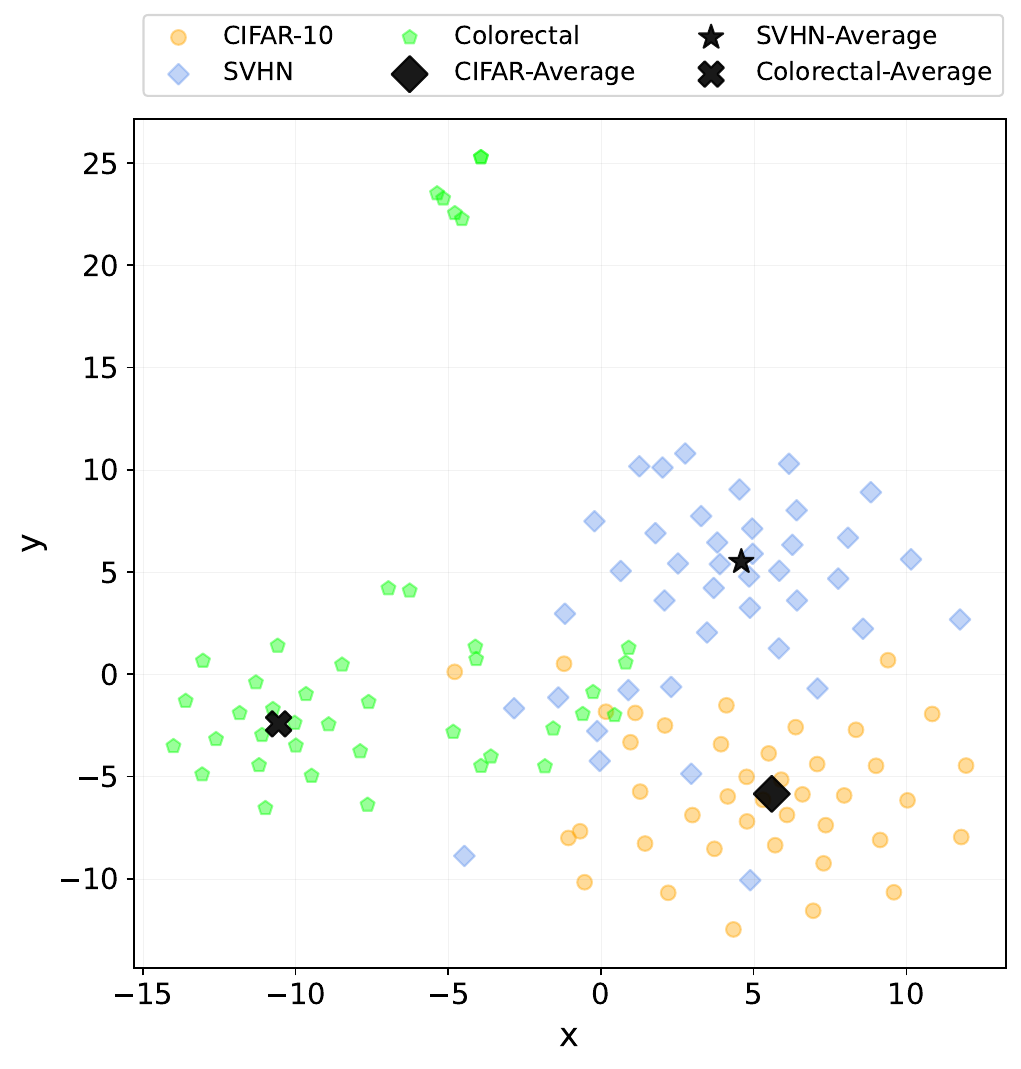}}}%
    \qquad
    \subfloat[SCA scores\centering]{{\includegraphics[scale=0.75,trim={0.5cm 0 0 0.5cm}]{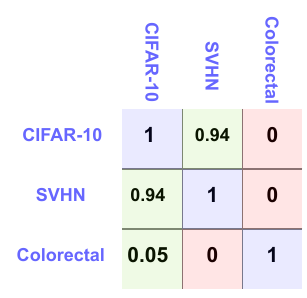} }}%
    
    \caption{\textbf{(a)} t-SNE representations of sample-wise converged \emph{Resnet-50} models for \emph{CIFAR-10}, \emph{street view house number (SVHN)} and \emph{colorectal histology} datasets. \textbf{(b)} SCA task affinities computed based on the density of task-specific sample-wise optima.}%
    \label{fig:tsne}%
\end{figure}

\subsubsection{Inferring inter-task similarities}
Equation \ref{eq:last} not only establishes an upper bound on the average training loss for dataset $\mathcal{D}$ in an MTL setup but also provides insights into inferring inter-task affinities. A higher density of task-specific sample-wise optima $\{\theta_i^{t\star}\}_{t=1}^T$ (lower $\psi_{\theta_i}$) implies a lower average training loss for the $i$-th sample (Theorem \ref{th:2}) and, by extension, for dataset $\mathcal{D}$. As discussed in Section \ref{sec:introduction}, two tasks are considered similar if a model can be trained for both tasks simultaneously (joint optimisation). Lower training loss serves as an indicator of the ease of joint optimisation. Therefore, $\psi_{\theta_i}$ can be leveraged to compute inter-task affinities. The average $\ell_1$-distance between the sample-wise optima of two tasks across $n$ samples offers an intuitive estimation of the affinity between those tasks, expressed as
\begin{equation}
    a_{\tau_i,\tau_j}= \frac{1}{n} \sum_{i=1}^n\lVert \theta^{\tau_i\star}_i - \theta^{\tau_j\star}_i \rVert_1, 
    \label{eq:pair}
\end{equation}

where $a_{\tau_i,\tau_j}$ is the affinity between tasks $\tau_i$ and $\tau_j$. Since the smoothness upper bound $\mathcal{H}$ (in Equation \ref{eq:dens}) is a constant value across all task pairs, the comparison between different pair-wise task affinities (required for task grouping) is only dependent on the distance between task-specific sample-wise optima.

\vspace{0.1cm}
Building on this, we introduce \emph{sample-wise convergence-based affinity} (SCA) scores to quantify task affinities. Given a set of $T$ tasks $\mathcal{T}=\{\tau_1,\tau_2,\ldots, \tau_T\}$ and their corresponding sample-wise optima for $n$ samples $\mathcal{O}=\{\theta_i^{\tau_1\star},\theta_i^{\tau_2\star},\ldots, \theta_i^{\tau_n\star}\}_{i=1}^n$, we compute SCA matrix $\mathbf{A} \in \mathbb{R}^{T\times T}$. 
Here, $\mathbf{A}_{i,j}$ indicates the affinity between tasks $\tau_i$ and $\tau_j$ ($a_{\tau_i,\tau_j}$), computed using Equation \ref{eq:pair}. Furthermore, every $i$-th row of $\mathbf{A}$ i.e., $\mathbf{A}_{i,:} \in \mathbb{R}^T$, is normalised to obtain the final affinity matrix as
\begin{equation}
    \mathbf{A}_{i,j}= 1 - \left(\frac{\mathbf{A}_{i,j}}{\textsc{max}(\mathbf{A}_{i,:})} \right), \quad \forall \mathbf{A}_{i,j} \in  \mathbf{A}_{i,:},
\end{equation}
where $\mathbf{A}_{i,j}$ is $j$-th element of row $\mathbf{A}_{i,:}$.

\vspace{0.1cm}
Figure \ref{fig:tsne}~(a) depicts a t-SNE illustration of $50$ task-specific sample-wise optima obtained from three datasets: CIFAR-10 \cite{krizhevsky2009learning}, SVHN \cite{netzer2011reading}, and the \emph{colorectal cancer histology} dataset \cite{kather2016multi}. Upon observation, the average pairwise distance separating the sample-wise optima for CIFAR-10 and SVHN appears smaller than the distance separating the optima of colorectal dataset from both CIFAR-10 and SVHN. This observation aligns with the intuitive expectation that the task affinity between CIFAR-10 and SVHN—both comprising low-resolution natural images—should be higher than the affinity between colorectal dataset, a dataset of histopathology or tissue images, and CIFAR-10/SVHN. The SCA matrix (Fig.~\ref{fig:tsne}~(b)) computed for these datasets is also consistent with this expectation, indicating that CIFAR-10 and SVNH are more similar to each other and may better benefit from joint training.

\subsubsection{Identifying task groupings}

The SCA scores gauge the affinity between task pairs, focusing on quantifying lower-order interactions. However, identifying task groupings requires measuring high-order interactions between different tasks. To address this, the proposed framework models inter-task relationships as a graph $\mathcal{G}(V,E)$, where each task is a node $v_i \in V$. Nodes/tasks $i$ and $j$ are connected by a weighted edge $e_{i,j} \in E$, with edge weight $w_{i,j}$ determined by the SCA-based affinity between tasks $i$ and $j$. The neighbourhood $\mathcal{N}_{v_i}$ of node $v_i$ is defined as the set of all nodes connected to $v_i$ in $\mathcal{G}(V,E)$. Each node $v_i$ is represented by a feature vector $\mathbf{f}_i = \mathbf{A}_{i,:}$, i.e., the $i$-th row of the affinity matrix $\mathbf{A}$, quantifying the SCA scores of task $\tau_i$ with every task in $\mathcal{T}$. 

\vspace{0.1cm}
The graph $\mathcal{G}(V,E)$ modelling pair-wise task-relationships is processed by Graph Attention Networks (GATs) to learn node or task-specific embeddings that encapsulate higher-order interactions or patterns beyond the direct pairwise SCA scores. In GATs, the attention mechanism represents each node as a learned linear combination of feature representations of nodes in the neighbourhood. Given node $v_i$ with neighbourhood $\mathcal{N}_i$ and representation $\mathbf{f}_i$, the operations of a $l$-th graph attention layer can be documented as

\begin{gather}
\mathbf{z}^{(l)}_i=\mathbf{W}^{(l)}\mathbf{f}_i, \\
\mathbf{h}_i^{(l+1)}=\sigma\left(\sum_{j\in\mathcal{N}_i}\alpha_{ij}^{(l)}\mathbf{z}_j\right),
\end{gather}
where $\mathbf{W}^{(l)}$ is the learnable weight matrix at $l$-th layer, $\mathbf{h}_i^{(l+1)}$ is the node embedding that acts as input to the next layer and $\alpha_{ij}^{(l)}$ is the learned additive attention score between nodes $v_i$ and $v_j$, and is computed as

\begin{gather}
e_{ij}^{(l)}=\textsc{LeakyRELU}\left(\mathbf{a}^{\top} [\mathbf{z}^{(l)}_i || \mathbf{z}^{(l)}_j]\right), \\
\alpha_{ij}^{(l)}=\frac{\textrm{exp}(e_{ij}^{(l)})}{\sum_{k \in \mathcal{N}_i}\textrm{exp}(e_{ik}^{(l)})}.
\end{gather}

\begin{figure}[t]
    \centering

    \includegraphics[scale=0.7,trim={3cm 0.5cm 3.2cm 0cm}]{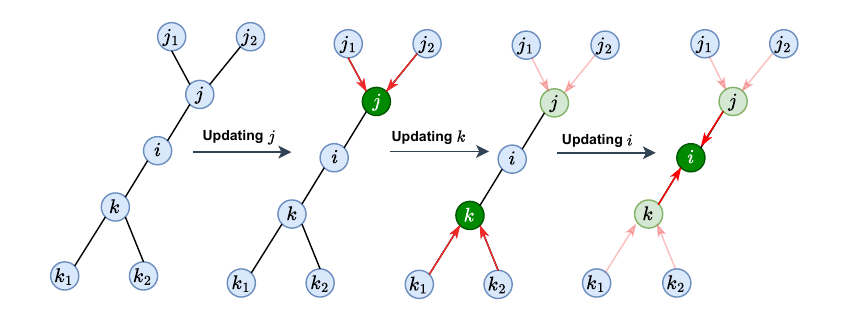}%

    \caption{An illustration of higher-order interactions captured by GATs. The representation learned for the $i$-th node depends not only on its immediate neighbours but also on the neighbourhoods of those neighbours. }%
    \label{fig:inter}%
\end{figure}

\begin{figure*}[t]
    \centering

    \includegraphics[scale=0.5,trim={1cm 0.5cm 1cm 0cm}]{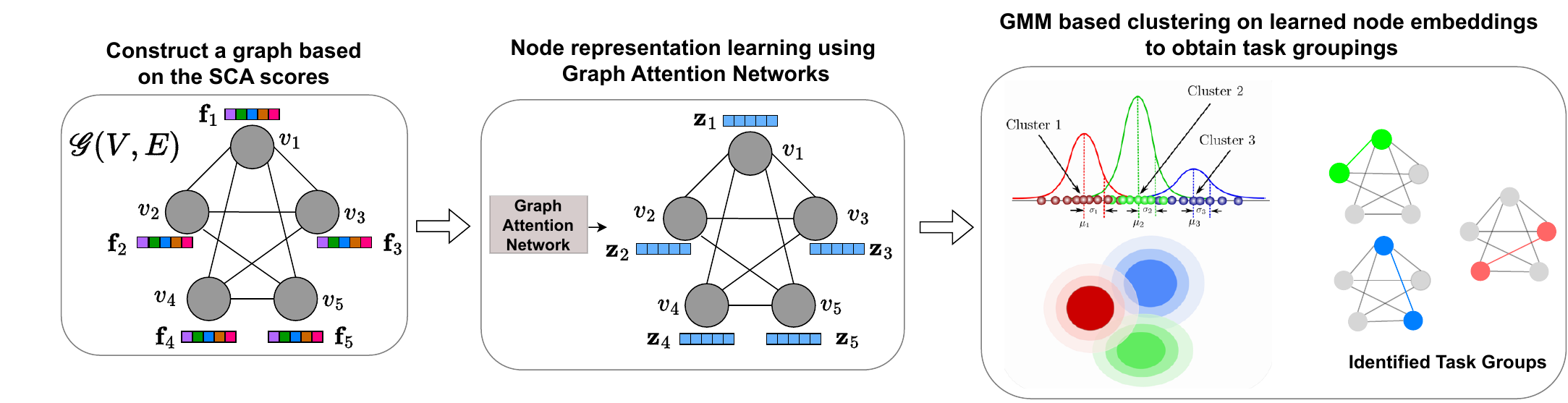}%

    \caption{Illustration of the process of obtaining task groupings from pair-wise SCA scores (affinity matrix) using graph attention network (GAT) and Gaussian mixture modelling.}%
    \label{fig:graphs}%
\end{figure*}

\noindent The representation $\mathbf{z}_i$ learned for a node $v_i$ by the GAT is a weighted combination of the representations of all nodes in its neighbourhood $\mathcal{N}_i$. Similarly, the representations of any two nodes, $v_j$ and $v_k$, within $\mathcal{N}_i$ depend on the representations of nodes in their respective neighbourhoods. Consequently, $\mathbf{z}_i$ indirectly captures information about the neighbourhoods of all nodes in $\mathcal{N}_i$, thereby modelling higher-order interactions among nodes or tasks. This behaviour is illustrated in Figure \ref{fig:inter}.

The proposed framework trains GAT in a self-supervised manner to reconstruct the input node representations $\mathbf{f}_i$ for each node $v_i$. The node embeddings $\mathbf{z}_i$ obtained after training for every node $v_i \in V$ are considered as a final node or task representations to learn task groupings. These node embeddings are clustered using Gaussian mixture models (GMM) with $b$ mixtures ($b$ is inference budget), and nodes or corresponding tasks in each mixture are considered as task grouping. Since GMM provides soft clustering, a node can be assigned to more than one cluster or grouping. To avoid singleton task groupings (as in our definition), a refinement step is carried out to assign any single node to the nearest or most similar cluster and hence, create a new cluster or corresponding task group. Figure \ref{fig:graphs} illustrates the overall process of obtaining task groupings from the affinity matrix $\mathbf{A}$.  

Details pertaining to the implementation of SCA scores and task groupings can be found in the supplementary document.

\begin{table}[t]
\centering
\caption{Characteristics of datasets used for experimentation.}
\label{tab:data}
\begin{sc}
\begin{small}
\resizebox{0.95\linewidth}{!}{
\begin{tabular}{c|c|c|c}
   \specialrule{.15em}{.15em}{.15em}

\textbf{\begin{tabular}[c]{@{}c@{}}Category\end{tabular}}                          & \textbf{Datasets}                                                               & \textbf{Tasks}                                                                & \textbf{\begin{tabular}[c]{@{}c@{}}\# Train, Test \& \\ Validation \\ Examples\end{tabular}} \\     \specialrule{.15em}{.15em}{.15em}

    \begin{tabular}[c]{@{}c@{}} Face Attribute \\Prediction\\ (Multi-tasking) \end{tabular} &  \begin{tabular}[c]{@{}c@{}}Celeb-A\end{tabular} & \begin{tabular}[c]{@{}c@{}} 9 classification \\ Tasks\end{tabular}            & \begin{tabular}[c]{@{}c@{}}162.7K, 19.9K
 \& \\ 19.9K\end{tabular}                                   \\  \specialrule{.1em}{.1em}{.1em}

        \begin{tabular}[c]{@{}c@{}} Molecular Property\\ Prediction\\ (Multi-tasking) \end{tabular} &  \begin{tabular}[c]{@{}c@{}}QM9\end{tabular} & \begin{tabular}[c]{@{}c@{}} 11 Regression\\ Tasks\end{tabular}            & \begin{tabular}[c]{@{}c@{}}70K, 40K \& \\ 20K\end{tabular}                                   \\

\specialrule{.1em}{.1em}{.1em}
\multirow{3}{*}{\begin{tabular}[c]{@{}c@{}}Critical \\ Patient Care\\ Tasks\end{tabular}} & \multirow{3}{*}{MIMIC-III}                                                      & \begin{tabular}[c]{@{}c@{}}In-hospital\\ Mortality \\ Prediction\end{tabular} & \begin{tabular}[c]{@{}c@{}}14698, 3222 \& \\ 3236\end{tabular}                               \\ \cmidrule{3-4} 
     &                                                                                 & \begin{tabular}[c]{@{}c@{}}Decompensation \\ Prediction\end{tabular}          & \begin{tabular}[c]{@{}c@{}}2388414, 520000 \& \\ 523208\end{tabular}                         \\ \cmidrule{3-4} 
         &                                                                                 & Phenotyping                                                                   & \begin{tabular}[c]{@{}c@{}}29280, 6371 \& \\ 6281\end{tabular}                               \\  \specialrule{.1em}{.1em}{.1em}
         
\multirow{4}{*}{\begin{tabular}[c]{@{}c@{}}Diagnosis \\ using \\ ECG signals\end{tabular}} & \multirow{4}{*}{\begin{tabular}[c]{@{}c@{}}MIMIC-III\\  Waveforms\end{tabular}} & \begin{tabular}[c]{@{}c@{}}Chronic \\ Kidney Disorder\end{tabular}            & \begin{tabular}[c]{@{}c@{}}382, 128 \& \\ 128\end{tabular}                                   \\ \cmidrule{3-4} 
 &                                                                                 & \begin{tabular}[c]{@{}c@{}}Conduction \\ Disorder\end{tabular}                & \begin{tabular}[c]{@{}c@{}}429, 143 \& \\ 144\end{tabular}                                   \\ \cmidrule{3-4}
&                                                                                 & \begin{tabular}[c]{@{}c@{}}Coronary \\ Atherosclerosis\end{tabular}           & \begin{tabular}[c]{@{}c@{}}1015, 339 \& \\ 339\end{tabular}                                  \\ \cmidrule{3-4}
&                                                                                 & Hypertension                                                                  & \begin{tabular}[c]{@{}c@{}}400, 134 \& \\ 134\end{tabular}                                   \\   \specialrule{.1em}{.1em}{.1em}
\multirow{3}{*}{Images}                                                                    & CIFAR-10                                                                        & Classification                                                                & \begin{tabular}[c]{@{}c@{}}37.5K, 12.5K \&\\  10K\end{tabular}                               \\ \cmidrule{2-4} 
      & SVHN-10                                                                         & \begin{tabular}[c]{@{}c@{}}Cropped Digit \\ Classification\end{tabular}       & \begin{tabular}[c]{@{}c@{}}58.7K, 14.7K \&\\  26K\end{tabular}                               \\ \cmidrule{2-4}
     & STL-10                                                                          & Classification                                                                & \begin{tabular}[c]{@{}c@{}}2.8K, 1K \&\\  1.2K\end{tabular}                                  \\ \cmidrule{2-4}
      & \begin{tabular}[c]{@{}c@{}}Colorectal \\  histology\end{tabular}                                                                            & \begin{tabular}[c]{@{}c@{}} 8-category\\ texture\\ classification \end{tabular}                                                                & \begin{tabular}[c]{@{}c@{}}2.8K, 1K \&\\  1.2K\end{tabular}                                    
                                      \\ \cmidrule{2-4}
      & \begin{tabular}[c]{@{}c@{}}Malaria \end{tabular}                                                                            & \begin{tabular}[c]{@{}c@{}} Malaria Detection\\ using \\Cell Images \end{tabular}                                                                & \begin{tabular}[c]{@{}c@{}}16.5K, 5.5K \&\\  5.5K\end{tabular}                                  \\    


        
      \specialrule{.15em}{.15em}{.15em}

\end{tabular}}
\end{small}
\end{sc}
\end{table}

\begin{figure*}[t]
    \centering
    \subfloat[\centering]{{\includegraphics[scale=0.28,trim={0 0.5cm 0 0.5cm}]{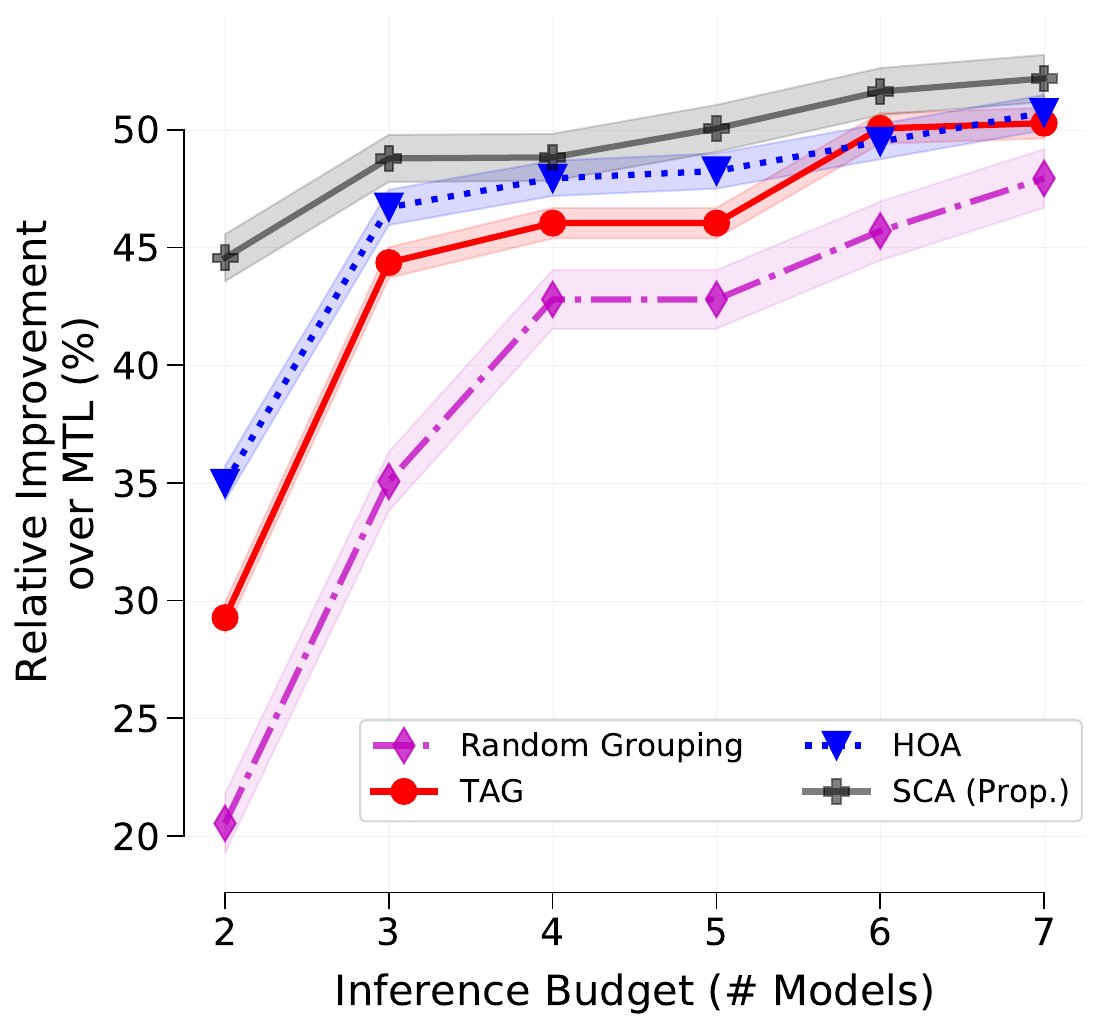} 
    \includegraphics[scale=0.28,trim={0 0.5cm 0 0.5cm}]{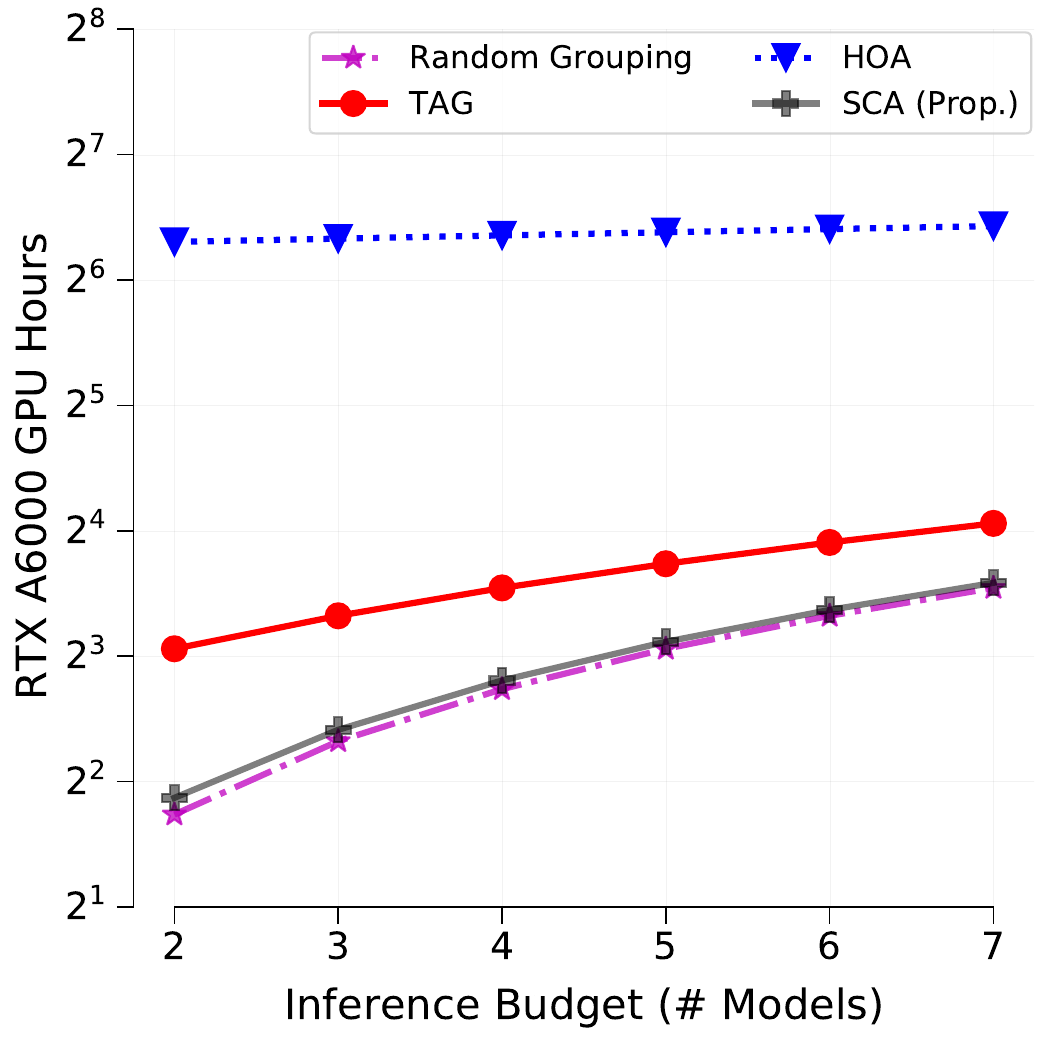}
    }}%
        \quad
    \subfloat[\centering]{{\includegraphics[scale=0.3,trim={0 0.5cm 0 0.5cm}]{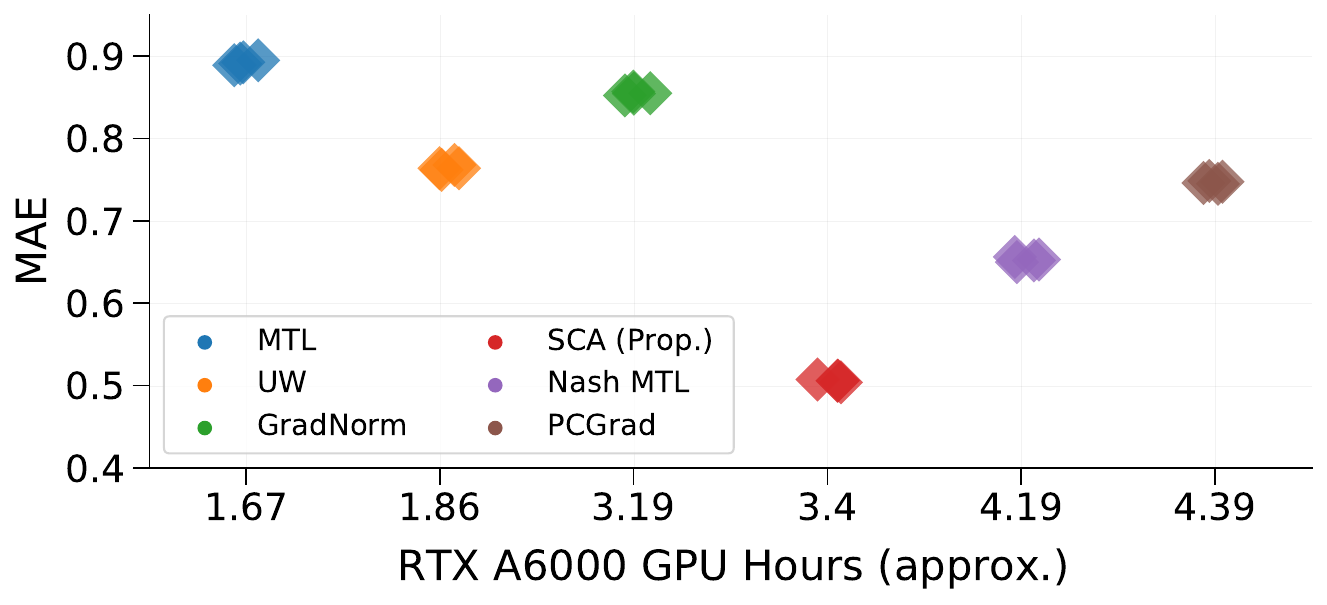} }}%
    
    \caption{Performance and computational efficiency of the proposed \emph{SCA}-based framework in comparison to prominent (a) task groupings and (b) multi-task learning (MTL) methods. For comparison with MTL methods, only \emph{SCA} with $2$ splits or task groupings is used.}%
    \label{fig:qm9_main}%
\end{figure*}

\section{Experiments}
\label{sec:exp}

\subsection{Datasets}
Table \ref{tab:data} documents the datasets and corresponding tasks used to evaluate the proposed framework. Each category refers to the set of tasks that are candidates for being trained simultaneously using a shared model either in a MTL or gradient-based meta-learning setup. The attributes of these categories are as follows:

\begin{itemize}[leftmargin=*]
\itemsep 0.3em

\item \textbf{Molecular property predictions:} This category deals with predicting $11$ different quantum chemical properties from the molecular structures represented as weighted graphs using Quantum Machine 9 (QM9) \cite{nandi2023multixc} dataset. The list of $11$ properties used in this work is provided in the supplementary document. 

\item \textbf{Face attribute predictions:} This category deals with predicting $9$ binary face attributes from celebrity images using CelebFaces Attributes Dataset (CelebA) \cite{liu2015deep}. The face attributes used in TAG \cite{fifty2021efficiently} are used here.

    \item \textbf{Patient care tasks:} These tasks involve predicting \emph{in-hospital mortality} based on the first 48 hours of ICU stay, \emph{decompensation} (i.e., decline in patient condition) within the next 24 hours, and identifying phenotypes based on the complete ICU stays using the MIMIC-III dataset \cite{johnson2016mimic,Hrayr_ML_17}. Mortality and decompensation predictions are binary classification tasks, while phenotyping involves assigning one or more predefined conditions from a list of 25 chronic and acute conditions to an input example \cite{Hrayr_ML_17}. Each example or ICU stay is represented by an hourly spaced time series where 76 clinical measurements are sampled at each time step. Except for mortality prediction, each ICU stay is represented by a variable-length time series.

    \item \textbf{Diagnosis using ECG signals:} In these prediction tasks, a one-minute long ECG signal, sampled from MIMIC-III waveform\footnote{\url{http://physionet.org/content/mimic3wdb/1.0/}} dataset \cite{johnson2016mimic}, is processed to diagnose chronic kidney disorder, conduction disorders, coronary atherosclerosis and hypertension. The ECG signals are pre-processed to obtain spectrograms, a time-frequency representation of signals, that are fed to models.   

    \item \textbf{Image classification:} CIFAR-10 \cite{krizhevsky2009learning}, STL-10 \cite{coates2011stl10}, street view house number (SVHN) \cite{netzer2011reading}, \emph{Colorectal histology} \cite{kather2016multi} and Malaria \cite{rajaraman2018pre} datasets are used for image classification tasks. CIFAR-10 and SVHN contain low-resolution images ($32\times32$), whereas STL-10 contains larger images ($96\times96$). On the other hand, \emph{Colorectal histology} and Malaria datasets contain histopathology and single cell images, respectively.
 
\end{itemize}

\begin{table*}[t]
    \caption{Performance of different task grouping methods on Celeb-A dataset in terms of (a) total absolute error and (b) average training time (RTX A6000 GPU hours).}
    \label{tab:celeb}
    \vspace{-0.25cm}
    \begin{minipage}{.5\linewidth}
      \caption*{(a) Mean Absolute Error}
      \centering
\resizebox{0.96\linewidth}{!}{
\begin{tabular}{c|cccc}
\toprule 
\textbf{Method $\downarrow$} & \multicolumn{4}{c}{\textbf{Splits}}                         \\ \toprule
\textbf{}       & \textbf{2}    & \textbf{3}   & \textbf{4}    & \textbf{5}   \\ \cmidrule{2-5}

Random                           & $50.1 \pm 0.085$  & $49.97 \pm 0.13$ & $49.75 \pm 0.08$  & $49.73 \pm 0.11$ \\
TAG                              & $49.66 \pm 0.095$ & $\textit{\textbf{49.55}} \pm \textit{\textbf{0.08}}$ & $\textit{\textbf{49.48}} \pm \textit{\textbf{0.066}}$ & $49.46 \pm 0.07$ \\
HOA                              & $49.76 \pm 0.08$  & $49.74 \pm 0.43$ & $49.72 \pm 0.1$   & $49.69 \pm 0.09$ \\

SCA (Prop.)                      & $\textit{\textbf{49.6}} \pm \textit{\textbf{0.24}}$   & $49.58 \pm 0.07$ & $\textit{\textbf{49.48}} \pm \textit{\textbf{0.05}}$  & $\textit{\textbf{49.44}} \pm \textit{\textbf{0.09}}$ \\ \bottomrule
\end{tabular}}
    \end{minipage}%
    \begin{minipage}{.5\linewidth}
      \centering
        \caption*{(b) Average Training Time }
\resizebox{0.96\linewidth}{!}{
\begin{tabular}{c|cccc}
\toprule 
\textbf{Method $\downarrow$} & \multicolumn{4}{c}{\textbf{Splits}}                         \\ \toprule
\textbf{}       & \textbf{2}    & \textbf{3}   & \textbf{4}    & \textbf{5}   \\  \cmidrule{2-5}

Random                           & $3.52 \pm 0.02$  & $ 5.24 \pm 0.02$ & $6.98 \pm 0.01$  & $8.62 \pm 0.02$ \\
TAG                              & $5.4 \pm 0.03$ & $7.17 \pm 0.04$ & $8.61 \pm 0.03$ & $10.45 \pm 0.02$ \\
HOA                              & $57.6 \pm 0.08$  & $ 59.5 \pm 0.06$ & $61.2 \pm 0.04$   & $62.7 \pm 0.06$ \\

SCA (Prop.)                      & $3.61 \pm 0.02$   & $5.31 \pm 0.03$ & $7.06\pm 0.02$  & $8.71 \pm 0.01$ \\ \bottomrule
\end{tabular}}
    \end{minipage} 
\end{table*}

\subsection{Models and training setup}

Molecular property and face attribute predictions are treated as MTL problems, employing hard parameter sharing in training multi-tasking models. In the QM9 dataset, a graph convolutional neural network\footnote{Multi-tasking variant of the model documented in \url{https://github.com/pyg-team/pytorch_geometric/blob/master/examples/qm9_nn_conv.py}} is utilised, with all layers shared except the last (prediction layer) across all $11$ regression tasks. This model is trained using mean square error as the loss function. For the CelebA dataset, the ResNet-15 based multi-tasking model from TAG \cite{fifty2021efficiently} is employed in this study.

\vspace{0.1cm}

For patient care, ECG prediction, and image classification categories, we conduct joint model training using Reptile, a first-order gradient-based meta-learning algorithm as detailed in \cite{thakur2021dynamic}. Each model is composed of shared layers along with task- or dataset-specific layers, where only the shared layers undergo joint training. For patient-care tasks, we implement the LSTM-based model used in \cite{thakur2021dynamic}. For ECG-based diagnosis tasks, a convolutional recurrent neural network with multiple output layers is employed. Additionally, ResNet-50, initialised with ImageNet weights, serves as the model for image classification. 
\vspace{0.1cm}

\subsection{Comparative methods}
The performance of the proposed framework is compared to random groupings, higher order approximations (HOA) \cite{standley2020tasks}, and task affinity groupings (TAG) \cite{fifty2021efficiently}. In random groupings, tasks are simply divided into $b$ groups, which represent either the inference budget or the allowable number of models. Details regarding TAG and HOA can be found in Section \ref{sec:back}. For each task category, we identify $k\leq b$ task groups using all the comparative methods and train a corresponding model for each task group. In instances where a task pertains to multiple groupings or is addressed by multiple models, the model delivering the best validation score is utilised for the final evaluation. The evaluation metrics encompass the average GPU training time and the average performance across tasks in all groupings.
\vspace{0.1cm} 

For multi-tasking setup, we also compare the performance of the proposed framework against prominent MTL methods that train a single model for all tasks but incorporate mechanisms to avoid gradient conflicts or negative transfer. These methods include GradNorm \cite{chen2018gradnorm}, Uncertainty Weighting (UW) \cite{kendall2018multi}, projecting conflicting gradients (PCGrad) \cite{yu2020gradient}, Nash equilibrium solution based MTL (Nash MTL) \cite{navon2022multi} and standard MTL (without any additional constraints to avoid gradient conflicts). Again, these methods have been discussed in Section \ref{sec:back}.  

\vspace{0.1cm}
Instead of using all available examples, the proposed framework samples only $N=100$ examples per dataset to perform sample-wise analysis and compute task or dataset affinities. The relative task affinities are found to captured effectively using the smaller number of samples, as outlined in Section \ref{ssec:sample_num}. More details regarding training setups, model architectures and hyperparameter settings can be found in the supplementary document.

\section{Results and discussion}

\subsection{Evaluation in multi-tasking setup}

\noindent \textsc{\textbf{Molecular property prediction:}} Figure \ref{fig:qm9_main} (a) illustrates the performance of the proposed framework compared to the prominent task grouping methods. All methods are evaluated with an inference budget of $2$ to $7$ models i.e.~$11$ regression tasks are split into $2$ to $7$ task groupings. The analysis of this figure highlights that the proposed framework significantly improves upon the standard multi-tasking model in terms of the average mean absolute error (MAE) across all $11$ tasks. Furthermore, the proposed framework either outperforms or shows comparable performance to HOA, TAG, and random splits across all settings. This indicates that the proposed framework is adept at identifying task groups that either mitigate negative transfer or facilitate information transfer among individual tasks. As the number of task groupings increases, the performance differences among all methods diminish. More task groupings inherently decrease the likelihood of negative transfer or gradient conflicts, resulting in similar performance. However, it is worth noting that the performance of ``informed'' task grouping methods, such as the proposed framework, TAG, and HOA, remains noticeably superior to random groupings across all settings.

This figure further illustrates that the average GPU running time (identifying task groupings and training models for each group) of the proposed framework is significantly less than TAG and HOA. As discussed earlier, it is expected as the proposed framework does not perform any ``complete'' model training to identify task groups as required in TAG and HOA. Moreover, the average running time of the proposed framework is slightly more than random groupings or splits.

The performance of the proposed framework with two task groupings, as illustrated in Figure \ref{fig:qm9_main} (b), is compared against prominent multi-tasking methods. The limitation of the number of task groupings to two is intentional to minimise the average running time. While complex multi-tasking methods outperform standard multitasking (MTL), the proposed framework demonstrates a significant performance improvement over these methods. Specifically, the proposed framework exhibits relative improvements of 45\% and 23.1\% compared to MTL and Nash MTL (best-performing baseline), respectively. Despite the computational efforts required for task grouping identification and training two multi-tasking models, the average running time of the proposed framework remains lower than that of PCGrad and Nash MTL.
\vspace{0.25cm}



\begin{figure}[t]
    \centering

    \includegraphics[scale=0.35,trim={1cm 0.5cm 0 0.5cm}]{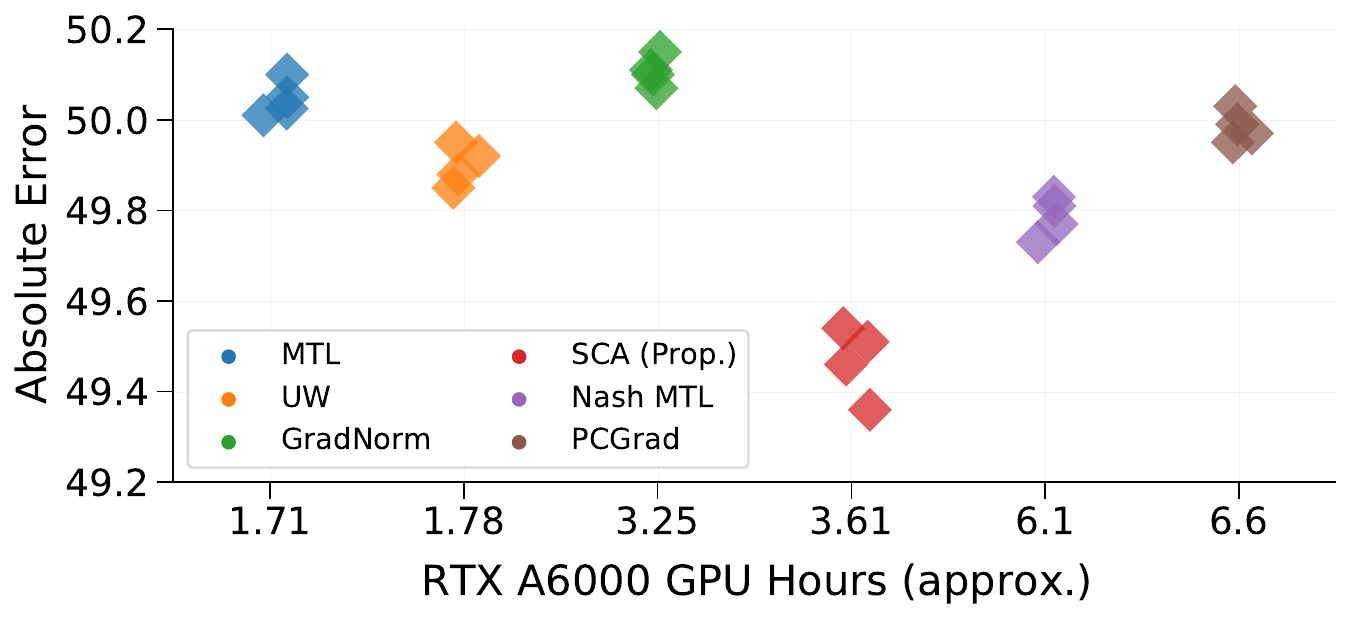}%

    \caption{Performance and computational efficiency of the proposed \emph{SCA}-based framework in comparison to prominent multi-task learning (MTL) methods on Celeb-A dataset.}%
    \label{fig:celeb_run}%
\end{figure}

\begin{figure}[t]
    \centering
    \centering
    \subfloat[Image classification\centering]{{\includegraphics[scale=0.23,trim={1cm 0.5cm 0 0.5cm}]{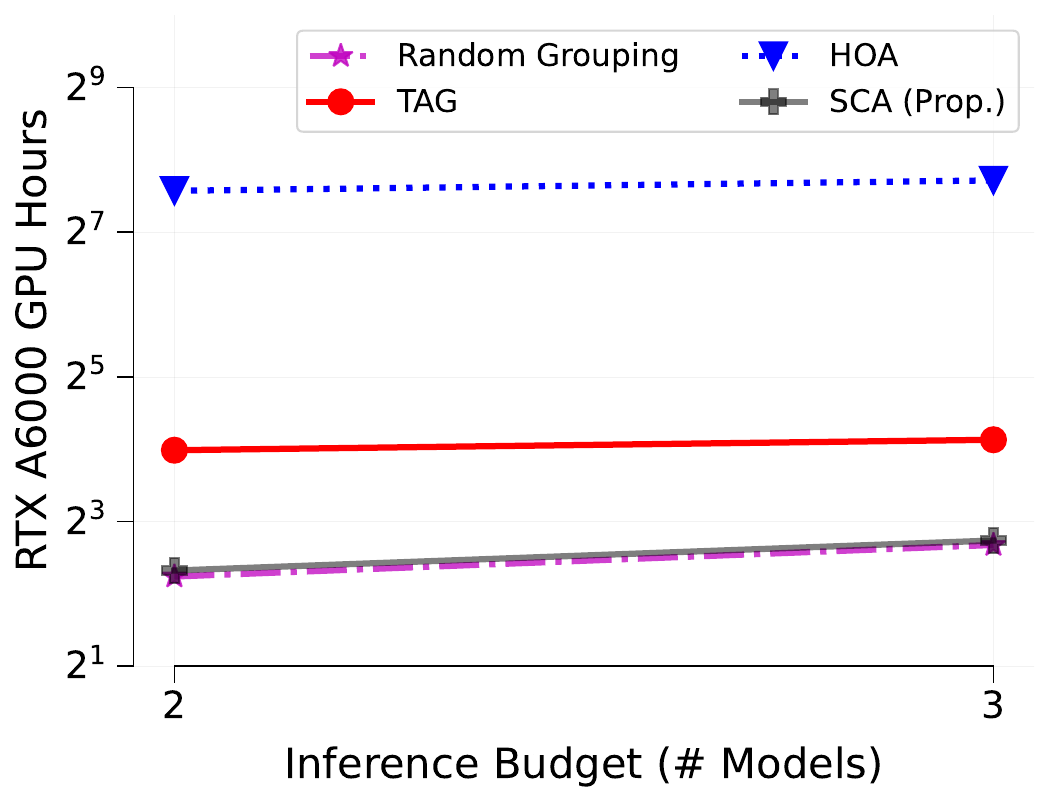}}}%
    \quad
    \subfloat[MIMIC-III Waveform\centering]{{\includegraphics[scale=0.23,trim={1cm 0.5cm 0 0.5cm}]{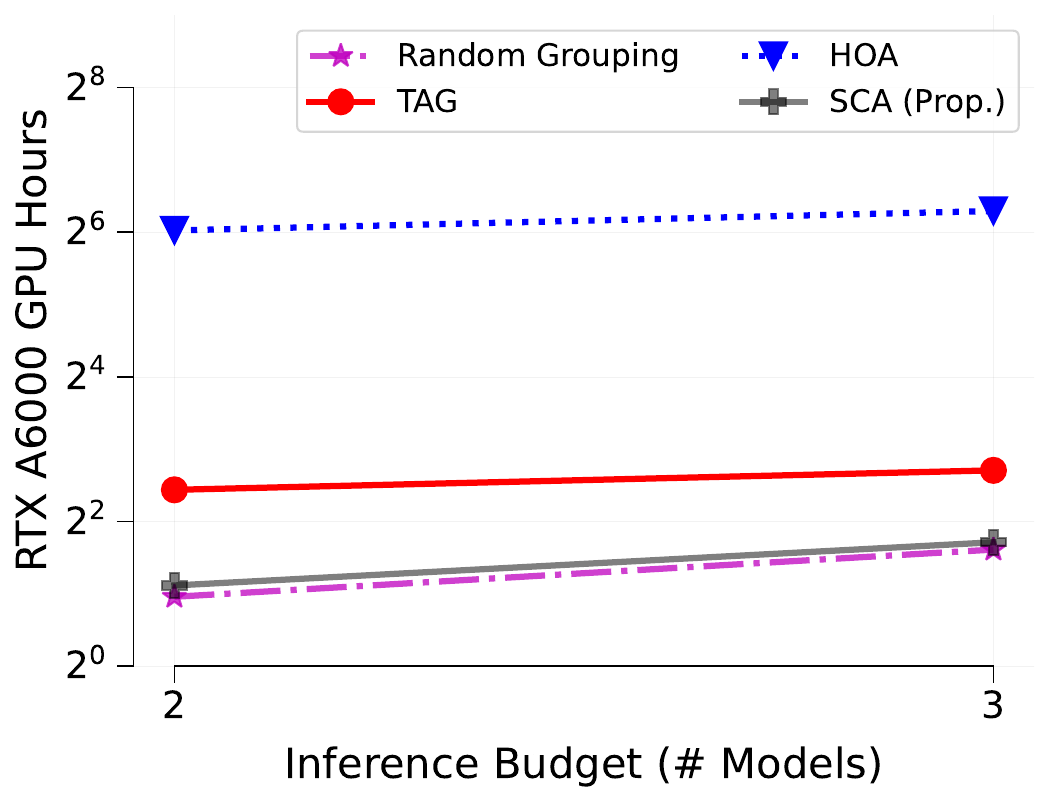}}}%

    \caption{Running time of different task grouping methods on \textbf{(a)} \emph{image classification} and \textbf{(b)} \emph{MIMIC-III Waveform} tasks.}%
    \label{fig:reptile_run}%
\end{figure}

\begin{table*}[t]
\centering
\caption{Performance of different task grouping methods on image classification tasks trained using \emph{Reptile}, a model-agnostic meta-learning algorithm. Balanced accuracy is used as the performance metric across all tasks.}
\label{tab:image}
\begin{sc}
\resizebox{0.9999\linewidth}{!}{
\begin{tabular}{ccccccclcccccc}

\toprule
\multicolumn{1}{c}{}                                               & \multicolumn{6}{c}{Splits-2}                                                                                                                                  & \multicolumn{1}{l}{} & \multicolumn{6}{c}{Splits-3}                                                                                                                                 \\ \cmidrule{2-7} \cmidrule{9-14} 
\textbf{METHODS $\Big \downarrow$}                                                            & \textbf{CIFAR} & \textbf{SVHN}  & \textbf{\begin{tabular}[c]{@{}c@{}}Colorectal\\ Cancer\end{tabular}} & \textbf{STL-10} & \textbf{Malaria} & \textit{Average} &                       & \textbf{CIFAR} & \textbf{SVHN} & \textbf{\begin{tabular}[c]{@{}c@{}}Colorectal\\ Cancer\end{tabular}} & \textbf{STL-10} & \textbf{Malaria} & \textit{Average} \\ \cmidrule{2-7} \cmidrule{9-14} 

\begin{tabular}[c]{@{}c@{}}Random\end{tabular}          & $78.25 \pm 0.92$          & $90.1\pm 0.95$           & $84.57 \pm 1.1$                                                                & $76.26 \pm 0.79$           & $96.25 \pm 0.75$            & $85.05 \pm 0.9$            &                       & $78.69 \pm 0.61$          & $90.28 \pm 0.48$         & $84.97 \pm 0.7$                                                                & $76.76 \pm 0.96$           & $96.12 \pm 0.39$            & $85.36 \pm 0.63$            \\
\begin{tabular}[c]{@{}c@{}}TAG\end{tabular}   & $\textit{\textbf{79.95}} \pm \textit{\textbf{0.45}}$ & $90.25 \pm 0.31$          & $85.15 \pm 0.47$                                                                & $77.2 \pm 0.54$            & $\textit{\textbf{96.94}} \pm \textit{\textbf{0.38}}$   & $85.9 \pm 0.43$             &                       & $\textit{\textbf{79.95}} \pm \textit{\textbf{0.45}}$          & $90.7 \pm 0.42$          & $85.15 \pm 0.47$                                                                & $76.4 \pm 0.61$            & $\textit{\textbf{96.94}} \pm \textit{\textbf{0.38}}$            & $85.83 \pm 0.47$            \\
\begin{tabular}[c]{@{}c@{}}HOA\end{tabular} & $79.3 \pm 0.43$           & $90.4 \pm 0.56$          & $85.45 \pm 0.39$                                                                & $77.32 \pm 0.41$           & $96.85 \pm 0.34$            & $85.86 \pm 0.42$           &                       & $79.74 \pm 0.57$          & $90.93 \pm 0.47$         & $ \textit{\textbf{85.64}} \pm \textit{\textbf{0.58}}$                                                                & $77.32 \pm 1.15$           & $96.85\pm 0.64$            & $86.1 \pm 0.68$             \\
SCA (Prop.)                                                          & $79.48 \pm 0.39$          & $\textit{\textbf{90.97}} \pm \textit{\textbf{0.41}}$ & $\textit{\textbf{86.27}}\pm\textit{\textbf{0.52}}$                                                       & $77.27 \pm 0.37$           & $96.86 \pm 0.29$            & $\textit{\textbf{86.17}} \pm \textit{\textbf{0.4}}$   &                       & $79.85 \pm 0.49$          & $\textit{\textbf{91.5}} \pm \textit{\textbf{0.52}}$          & $85.43 \pm 0.39$                                                                & $77.1 \pm 0.53$            & $\textit{\textbf{97.05}} \pm \textit{\textbf{0.31}}$   & $\textit{\textbf{86.2}} \pm\textit{\textbf{0.45}}$         
\\
\bottomrule   
\end{tabular}}
\end{sc}
\end{table*}

\begin{table*}[t]
\centering
\caption{Performance of different task grouping methods on MIMIC-III Waveform prediction tasks. Area under the ROC is used as the performance metric across all tasks.}
\label{tab:ecg}
\resizebox{0.95\linewidth}{!}{
\begin{tabular}{c|ccccccccc}
\toprule
\multirow{2}{*}{\textbf{TASKS} $\Big \downarrow$ \;\;\textbf{METHODS} $\rightarrow$}

& \multicolumn{4}{c}{Splits-2}         &  & \multicolumn{4}{c}{Splits-3}   \\  \cmidrule{2-5} \cmidrule{7-10} 
                                                                   & \textbf{Random} & \textbf{TAG}   & \textbf{HOA}   & \textbf{SCA (Prop.)} &  & \textbf{Random} & \textbf{TAG}   & \textbf{HOA}   & \textbf{SCA (Prop.)}   \\  \toprule
\begin{tabular}[c]{@{}c@{}}Chronic Kidney Disorder\end{tabular}  & $0.572 \pm 0.015$  & $0.615 \pm 0.009$ & $\textit{\textbf{0.618}} \pm \textit{\textbf{0.007}}$ & $0.611 \pm 0.011$       &  & $0.592 \pm 0.012$  & $0.627 \pm 0.006$ & $\textit{\textbf{0.631}} \pm \textit{\textbf{0.007}}$ & $0.629 \pm 0.01$ \\
\begin{tabular}[c]{@{}c@{}}Conduction  Disorder\end{tabular}     & $0.661 \pm 0.017$  & $0.687 \pm 0.011$ & $0.685 \pm 0.014$ & $\textit{\textbf{0.693}} \pm \textit{\textbf{0.013}}$      &  & $0.671 \pm 0.012$  & $0.689 \pm 0.008$ & $0.687 \pm 0.011$ & $\textit{\textbf{0.691}} \pm \textit{\textbf{0.009}}$ \\
\begin{tabular}[c]{@{}c@{}}Coronary Atherosclerosis\end{tabular} & $0.636 \pm 0.013$  & $0.649 \pm 0.01$ & $0.652 \pm 0.008$ & $\textit{\textbf{0.668}} \pm \textit{\textbf{0.007}}$       &  & $0.659 \pm 0.014$  & $0.659 \pm 0.013$ & $0.652 \pm 0.016$ & $\textit{\textbf{0.668}} \pm \textit{\textbf{0.007}}$  \\
Hypertension                                                       & $0.612 \pm 0.016$  & $0.622 \pm 0.01$ & $0.619 \pm 0.007$ & $\textit{\textbf{0.628}} \pm \textit{\textbf{0.014}}$       &  & $0.634 \pm 0.011$  & $\textit{\textbf{0.654}} \pm \textit{\textbf{0.008}}$ & $0.634 \pm 0.01$ & $0.639 \pm 0.01$ \\

\emph{Average}  & $0.62 \pm 0.01$  & $0.643 \pm 0.01$  & $0.644 \pm 0.009$ & $\textit{\textbf{0.65}} \pm \textit{\textbf{0.011}}$       &  & $0.639 \pm 0.012$  & $\textit{\textbf{0.657}} \pm \textit{\textbf{0.009}}$ & $0.651 \pm 0.011$ & $\textit{\textbf{0.657}} \pm \textit{\textbf{0.01}}$ \\ \bottomrule
\end{tabular}}
\end{table*}

\begin{table}[t]
\centering
\caption{Performance of different task grouping methods on patient care tasks trained using \emph{Reptile}, a model-agnostic meta-learning algorithm. AUROC is used as the performance metric across all tasks.}
\label{tab:mimic}
\begin{sc}
\resizebox{0.9999\linewidth}{!}{
\begin{tabular}{c|c|c|c|c}
\toprule
\textbf{TASKS}                      & \textbf{Random} & \textbf{HOA}   & \textbf{TAG}   & \textbf{SCA (prop.)} \\ \toprule
Mortality  & $0.842\pm 0.002$   & $\textbf{0.845}\pm \textbf{0.001}$ & $\textbf{0.845}\pm \textbf{0.001}$ &  $\textbf{0.845}\pm \textbf{0.001}$     \\ 
Decompensation       & $0.869\pm 0.003$  & $\textbf{0.874}\pm\textbf{0.002}$ & $\textbf{0.874}\pm\textbf{0.002}$ & $\textbf{0.874}\pm\textbf{0.002}$     \\ 
Phenotyping          & $0.762\pm0.001$  & $\textbf{0.767}\pm\textbf{0.001}$ &  $\textbf{0.767}\pm\textbf{0.001}$ &  $\textbf{0.767}\pm\textbf{0.001}$     \\ 
Average              & $0.824\pm0.002$  &$\textbf{0.845}\pm \textbf{0.001}$ &$\textbf{0.845}\pm \textbf{0.001}$ &$\textbf{0.845}\pm \textbf{0.001}$      \\ \bottomrule

\end{tabular}}
\end{sc}
\end{table}

\noindent \textsc{\textbf{Face attribute prediction:}} Table \ref{tab:celeb} provides a detailed overview of the performance of various task grouping methods on the Celeb-A dataset. HOA, TAG, and the proposed framework exhibit comparable performance across all settings. However, the inherently lower computational requirements of the proposed framework still make it a more desirable solution. Furthermore, Figure \ref{fig:celeb_run} illustrates the performance of the proposed framework with two splits or task groupings against multi-tasking baselines. Once again, the average performance of the proposed framework is either superior or comparable, underscoring its effectiveness in identifying relevant task groupings. Moreover, the average training time of the proposed framework is approximately twice that of MTL but still less than that of Nash MTL (the best-performing baseline).


\subsection{Evaluation in meta-learning setup}

\noindent \textsc{\textbf{Image Classification:}} We consider each image dataset as a separate task, and the aim is to train shared models with task or dataset-specific layers. Table \ref{tab:image} documents the performance of different task grouping methods for the image classification category. The analysis of this table highlights that all task grouping methods perform significantly better than random task groupings. This shows that meta-learning based joint training can also benefit from informed task grouping approaches. The proposed framework (SCA) exhibits comparable or better performance that HOA and TAG for both $2$ and $3$ splits. Hence, it can be inferred that the proposed method can identify effective task groupings as TAG and HOA while requiring only a fraction of the computational resources (Figure \ref{fig:reptile_run} (a)).

\vspace{0.1cm}
\noindent \textsc{\textbf{Diagnosis using ECG signals:}}  The performance trends documented in Table \ref{tab:ecg} for this category are similar to image classification.  The proposed framework (SCA) significantly outperforms random groupings for both $2$ and $3$ splits. Specifically, for $2$ splits, the proposed framework demonstrates superior performance compared to HOA and TAG, with an average AUC of $0.65 \pm 0.011$, surpassing HOA ($0.644 \pm 0.009$) and TAG ($0.643 \pm 0.01$). For $3$ splits, the proposed framework achieves an average AUC of $0.657 \pm 0.01$, matching TAG while requiring fewer computational resources, as shown in Figure \ref{fig:reptile_run} (b). 

\vspace{0.1cm}
\noindent \textsc{\textbf{Patient care tasks:}} Table \ref{tab:mimic} shows that the performance of HOA, TAG and the proposed framework for obtaining $2$ task groupings from $3$ patient care tasks. HOA, TAG and the proposed framework resulted in same task groupings and hence exhibiting similar performance. On the other hand, random groupings exhibited slightly lower performance than the other methods.

\subsection{Training dynamics and SCA scores}

\begin{figure}[t]
    \centering
    \subfloat[\centering]{{\includegraphics[scale=0.2725,trim={0 0.5cm 0 0.5cm}]{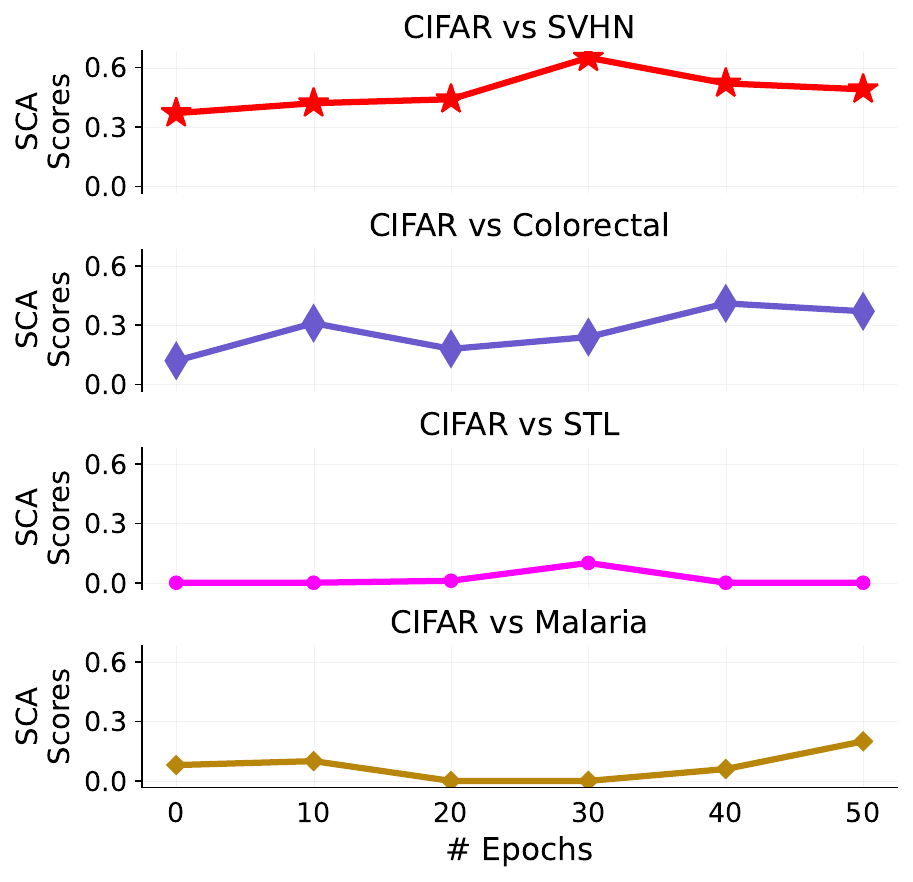}}}%
     \quad
     \subfloat[\centering]{{\includegraphics[scale=0.2725,trim={0 0.5cm 0 0.5cm}]{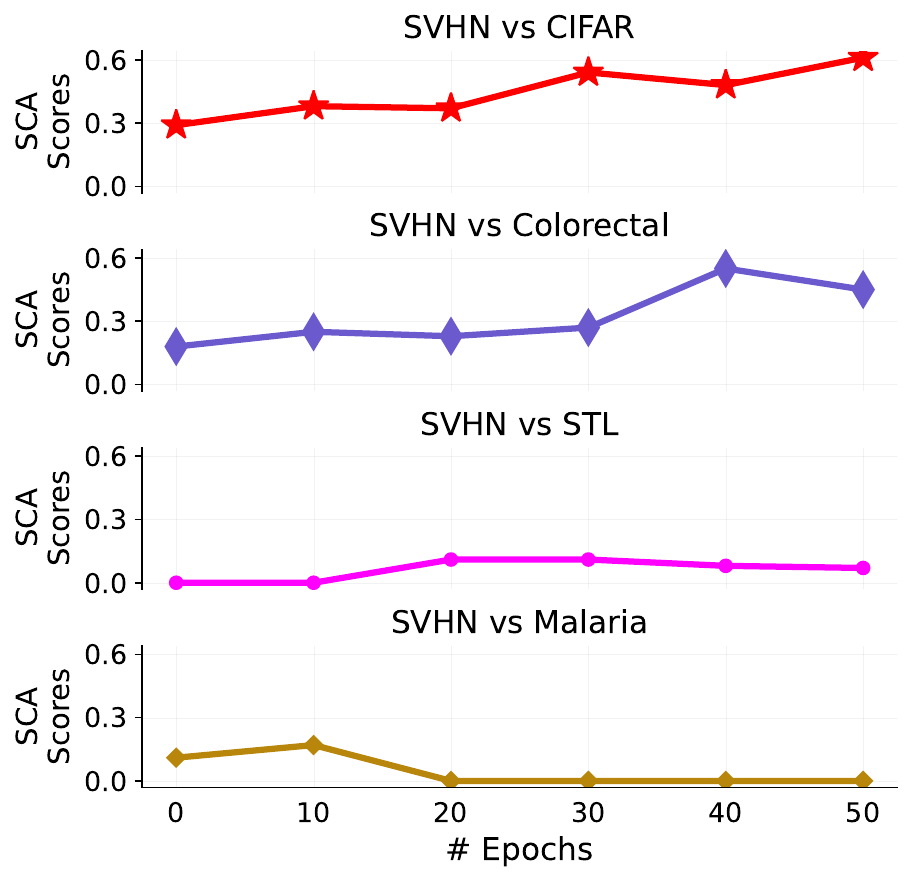} }}%
    \caption{Dynamics of the SCA scores as the joint training of all $5$ image datasets progresses.}%
    \label{fig:epochs}%
\end{figure}

\begin{figure}[t]
    \centering
    \subfloat[\centering]{{\includegraphics[scale=0.38,trim={0 0.5cm 0 0.5cm}]{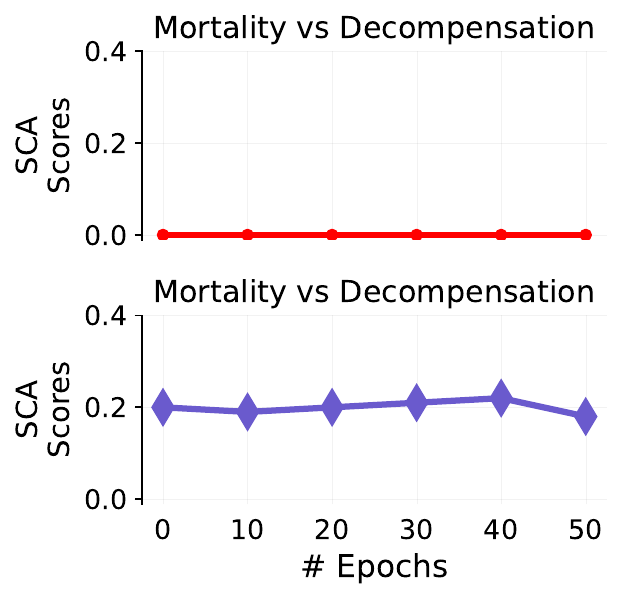}}}%
     \quad
     \subfloat[\centering]{{\includegraphics[scale=0.38,trim={0 0.5cm 0 0.5cm}]{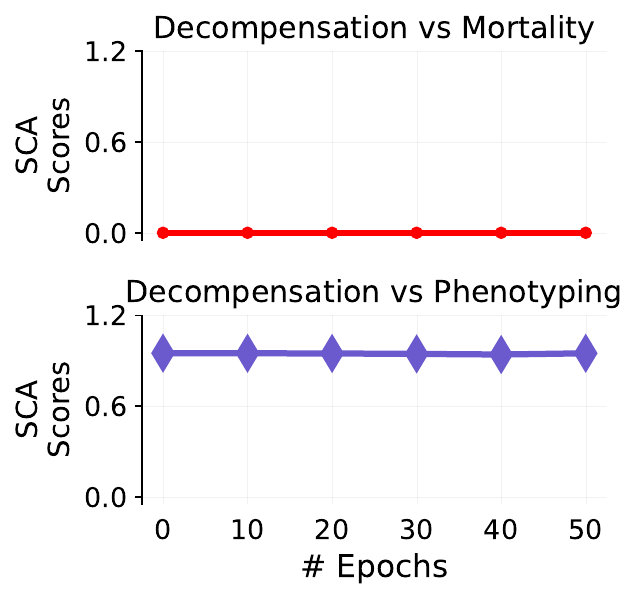} }}%
    \caption{Dynamics of the SCA scores as the joint training of all $3$ patient care tasks progresses.}%
    \label{fig:mimic_epochs}%
\end{figure}

The SCA scores are calculated based on the initial model state, without any training. It is essential to investigate how these scores evolve during the joint optimisation of multiple tasks. To explore this, we compute SCA scores at regular intervals during the joint training of image classification and patient care tasks in a meta-learning setup. Figures \ref{fig:epochs} and \ref{fig:mimic_epochs} illustrate the evolution of SCA scores for these respective categories. Our analysis reveals that while SCA scores fluctuate during training, the relative similarities among tasks remain largely consistent. For instance, in Figure \ref{fig:epochs} (a), the SCA score-based similarity of CIFAR with SVHN and the Colorectal Histology dataset is consistently higher than with STL-10 and Malaria. A similar pattern is observed in Figure \ref{fig:epochs} (b), where CIFAR and Colorectal Histology consistently show more similarity with SVHN than with other datasets. In patient care tasks, mortality prediction consistently aligns more closely with phenotyping, and decompensation prediction similarly aligns with phenotyping throughout training. This  analysis highlights that sample-wise similarities are largely preserved regardless of the model state, affirming the robustness of SCA scores in tracking task similarities during training.

\subsection{Influence of Sample Size on SCA Scores}
\label{ssec:sample_num}
\begin{figure}[t]
    \centering
    \subfloat[\centering]{{\includegraphics[scale=0.2725,trim={0 0.5cm 0 0.5cm}]{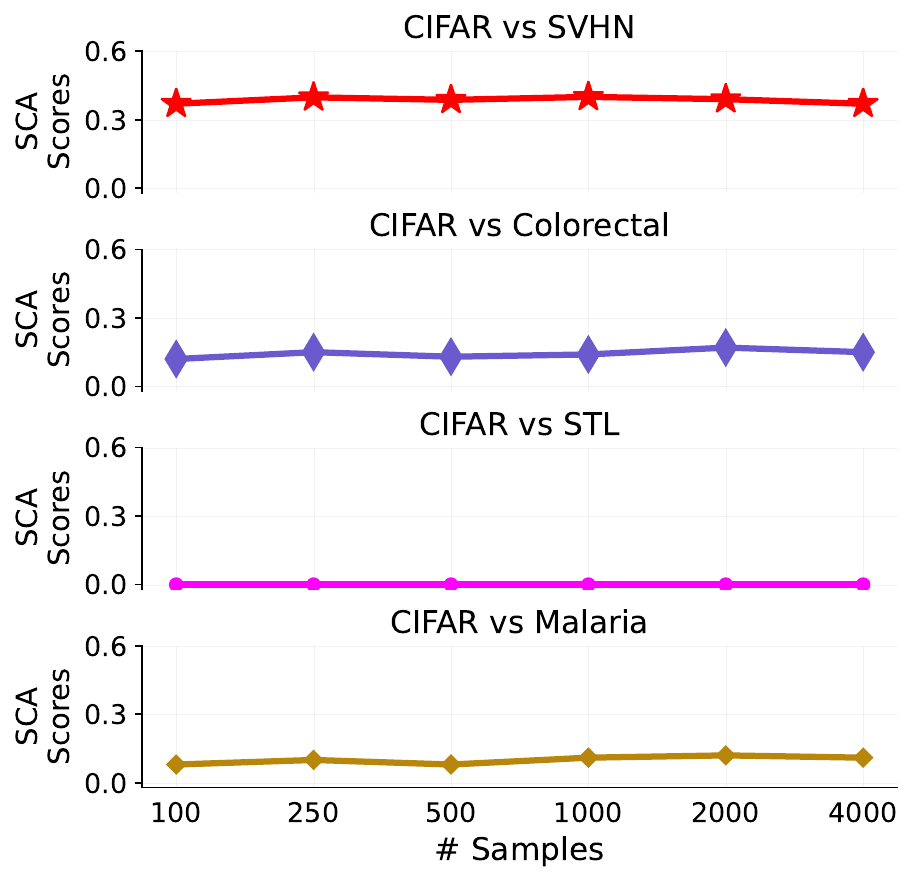}}}%
     \quad
     \subfloat[\centering]{{\includegraphics[scale=0.2725,trim={0 0.5cm 0 0.5cm}]{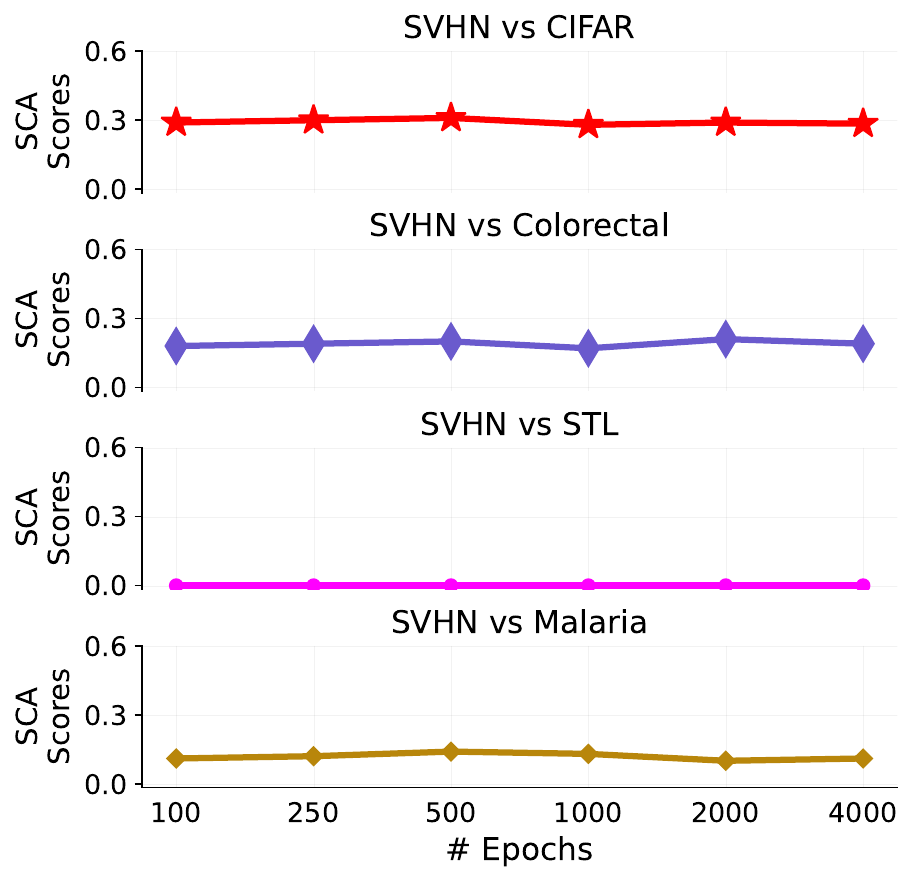} }}%
    \caption{Impact of sample size on SCA scores computed for (a) CIFAR amd (b) SVHN against other datasets.}%
    \label{fig:img_samples}%
\end{figure}

\begin{figure}[t]
    \centering
    \subfloat[\centering]{{\includegraphics[scale=0.38,trim={0 0.5cm 0 0.5cm}]{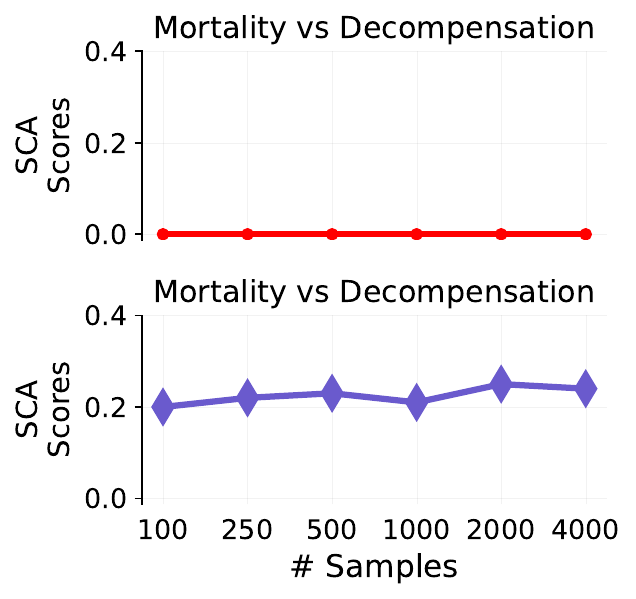}}}%
     \quad
     \subfloat[\centering]{{\includegraphics[scale=0.38,trim={0 0.5cm 0 0.5cm}]{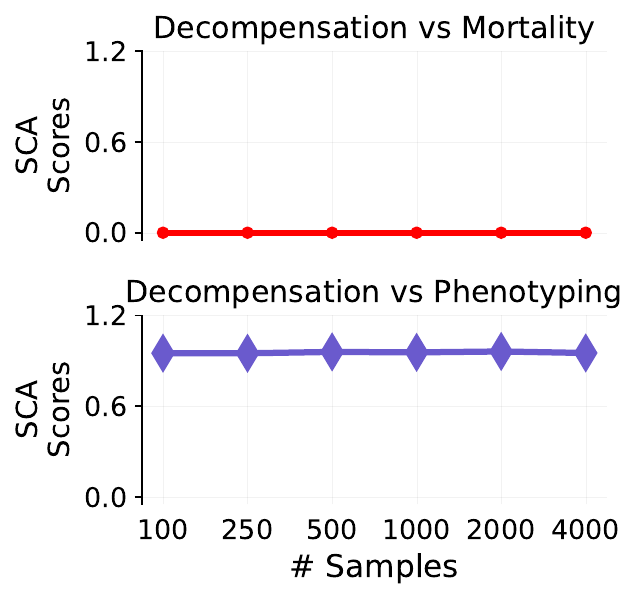} }}%
    \caption{Impact of sample size on SCA scores computed for (a) CIFAR and (b) SVHN against other datasets.}%
    \label{fig:mimic_samples}%
\end{figure}

As discussed in Section \ref{sec:exp}, we utilized only $100$ samples per task for the computation of SCA scores. In order to assess the impact of sample size on SCA scores, we systematically increased the number of samples to ${250, 500, 1000, 2000, 4000}$ and observed the resulting deviations in SCA scores for both image and patient care tasks. Figures \ref{fig:img_samples} and \ref{fig:mimic_samples} visually represent the changes in SCA scores corresponding to variations in sample size for image and patient care tasks, respectively. The analysis of these figures highlights that only marginal changes in SCA scores are discernible with the increment in sample size. These subtle variations are insufficient to induce any alterations in the relative task similarities captured by SCA scores. This observation suggests that even a modest number of samples is effective in discerning the inter-task relationships within the proposed framework.

\section{Conclusion}
Departing from conventional approaches, this paper expands the analysis of sample-wise convergence to encompass joint optimisation frameworks observed in multi-tasking and gradient-based meta-learning. Through a comprehensive theoretical examination, we unveil a significant relationship between sample-wise optima and the efficient optimisation of joint loss, providing insights to discern similarities among tasks engaged in joint training. Building on these insights, the paper introduces an efficient task grouping framework that employs sample-wise convergence analysis to calculate pairwise task affinities. Additionally, it integrates graph attention networks with a clustering mechanism to deduce higher-order task relationships, thereby facilitating effective task groupings. Rigorous experimentation across diverse datasets underscores the competitiveness of the proposed framework, achieving performance on par with state-of-the-art task grouping approaches while utilising a fraction of the computational resources.

Despite unearthing intriguing findings, it is essential to note that the task affinities computed in this work may or may not serve as a measure of semantic similarity among tasks. These affinities only illuminate the ease of joint optimisation for a subset of tasks. Given that joint optimisation is contingent on model architecture or complexity, it is plausible that SCA-based task affinities may vary with changes in the model architecture.

One potential limitation of this work is higher memory complexity, especially for very large models, required to compute pairwise distances among multiple task-specific sample-wise optima. However, as illustrated in Fig.~\ref{fig:img_samples}, only a handful of examples are enough to capture the average task affinity trends rendering these computational challenges negligible.

Instead of using randomly selected samples, future work will focus on selecting coresets or inducing points to compute SCA scores capturing a holistic view of data or task relationships.

\bibliographystyle{IEEEtran}
\bibliography{ref}

\begin{IEEEbiography}[{\includegraphics[width=1in,height=1.25in,clip,keepaspectratio]{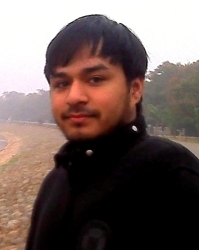}}]
{Dr. Anshul Thakur} 
received his PhD from the Indian Institute of Technology Mandi. He is currently a Departmental Lecturer in Clinical Machine Learning and is affiliated with the Computational Health Informatics Lab at the University of Oxford. His research interests centres around AI for healthcare, federated learning, multi task learning, data democratisation and AI4Science.  
\end{IEEEbiography}

\begin{IEEEbiography}[{\includegraphics[width=1in,height=1.25in,clip,keepaspectratio]{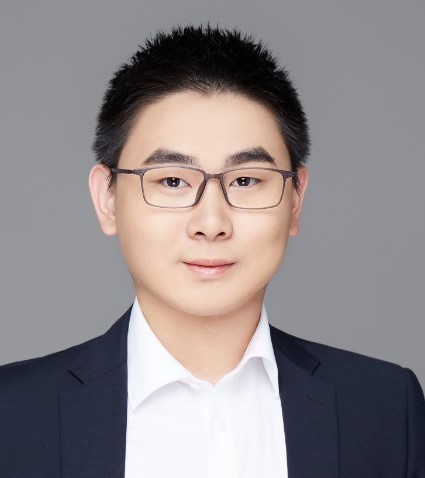}}]
{Mr. Yichen Huang} is a Master's degree student in the Department of Computer Science at Oxford University. He has received the prestigious Hoare Prize, awarded to the best-performing computer science student at Oxford. His research interests include machine learning, combinatorial optimisation, and algorithm analysis.

\end{IEEEbiography}

\begin{IEEEbiography}[{\includegraphics[width=1in,height=1.25in,clip,keepaspectratio]{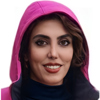}}]
{Dr. Soheila Molaei} obtained her PhD in Large-scale Graph Summarization from the University of Tehran in 2021, having completed her MSc at the same institution in 2016. Her research is primarily focused on deep graph neural networks, probabilistic link prediction, polymer informatics, drug discovery, and clinical AI. Presently, she is engaged in developing generative AI, including large language models (LLMs), to create new polymers that meet specific characteristics.
\end{IEEEbiography}

\begin{IEEEbiography}[{\includegraphics[width=1in,height=1.25in,clip,keepaspectratio]{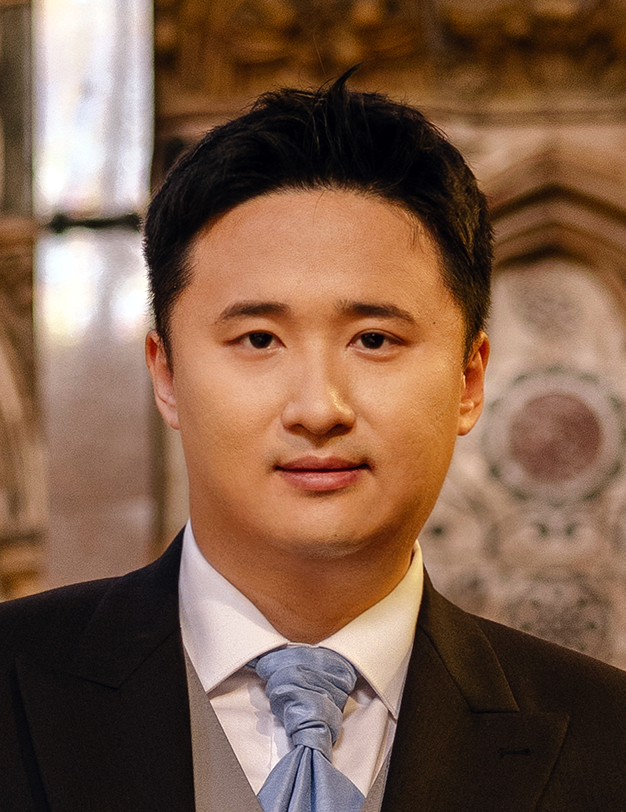}}]
{Dr. Yujiang Wang} 
received his PhD degree from Imperial College London, and currently, he is a Research Scientist and Associated Head of Machine Learning Lab at the Digital Health Group, Oxford Suzhou Centre for Advanced Research. 
He obtained a Bachelor's degree from Tsinghua University and was awarded two MSc with Distinction from University College London and Imperial College London, respectively. His research interest centres around AI for healthcare, AI4Science, computer vision, etc. 
\end{IEEEbiography}

\begin{IEEEbiography}  [{\includegraphics[width=1in,height=1.25in,clip,keepaspectratio]{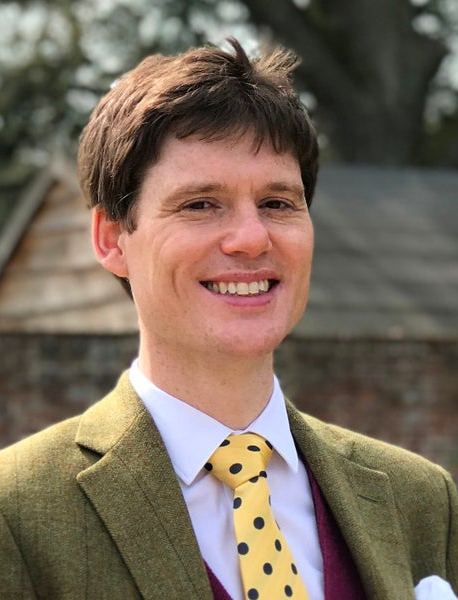}}]
{Prof. David A. Clifton} 
is the Royal Academy of Engineering Chair of Clinical Machine Learning at the University of Oxford and leads the Computational Health Informatics (CHI) Lab, which focuses on \say{AI for Healthcare}. He is also NIHR Research Professor appointed as the first non-medical scientist to the NIHR's \say{flagship chair}, a principal investigator of the Digital Health Group at Oxford Suzhou Centre for Advanced Research, a co-director of Oxford-CityU Centre for Cardiovascular Engineering in Hong Kong,  an investigator of the Pandemic Sciences Institute at the University of Oxford, a Fellow of the Alan Turing Institute, a Visiting Chair in AI for Health at the University of Manchester, and a Fellow of Fudan University, China. His research has won over 40 awards, including a Grand Challenge award from EPSRC, the Vice-Chancellor's Innovation Prize from the University of Oxford, the IEEE Early Career Award given to one engineer annually, etc. 
\end{IEEEbiography}

\vfill

\end{document}


\maketitle

\section{Dataset details}

\vspace{0.5cm}
\subsection*{List of tasks used for Celeb-A experiments}
\begin{multicols}{2}
\begin{enumerate}
\item $5$ o'clock  shadow
\item Black hair
\item Blond hair
\item Brown hair
\item Goatee
\item Moustache
\item No beard
\item Rosy cheeks
\item Wearing hat
\end{enumerate}
\end{multicols}

\subsection*{List of molecular properties/tasks used for QM9 experiments}
\begin{enumerate}
    \item $\mu$, Dipole moment
\item $\alpha$, Isotropic polarisability
\item $\epsilon_{\textrm{HOMO}}$, Highest unoccupied molecular orbital energy 
\item $\epsilon_{\textrm{LUMO}}$, Lowest unoccupied molecular orbital energy 
\item $\Delta \epsilon$, Gap between $\epsilon_{\textrm{HOMO}}$ and $\epsilon_{\textrm{LUMO}}$
\item $\langle R^2 \rangle$,  Electronic spatial extent
\item $\textrm{ZPVE}$,  Zero point vibrational energy
\item $c_{\textrm{v}}$, Heat capavity at 298.15K 
\item $U_0^{\textrm{ATOM}}$,  Atomization energy at 0K
\item $U^{\textrm{ATOM}}$,  Atomization energy at 298.15K
\item $H^{\textrm{ATOM}}$,  Atomization enthalpy at 298.15K
\item $G^{\textrm{ATOM}}$,  Atomization free energy at 298.15K
\end{enumerate}

\subsection*{Pre-processing ECG data}
MIMIC-III waveforms contain the ECG signals having a sampling rate of 256 Hz. A short-term Fourier transform is performed on each signal using a segment size of 64 samples and an overlap of 32 samples to obtain spectrograms (time-frequency representation). Spectrograms are normalised and converted to a log scale to obtain the final representation of an ECG signal.

\begin{table*}[t]
\centering 
\caption{Parameter setting used to train multi-tasking models for Celeb-A and QM9 datasets.}
\label{tab:mtl_pars}
\begin{small}
\begin{sc}

\begin{tabular}{ccccc}
\toprule
\textbf{Dataset} & \textbf{Batch Size} & \textbf{Optimiser} & \textbf{Learning Rate} & \textbf{Momentum} \\ \toprule
Celeb-A          & 256                 & SGD                & 0.005                  & 0.9               \\
QM9              & 128                 & SGD                & 0.001                  & 0.9   \\ \bottomrule          
\end{tabular}
\end{sc}
\end{small}
\end{table*}

\begin{table*}[t]
\centering
\caption{Parameters used to train models using Reptile.}
\label{tab:rep}
\begin{small}
\begin{sc}
\begin{tabular}{cccccc}
\toprule
\textbf{Task Category} & \textbf{Batch Size} & \textbf{\begin{tabular}[c]{@{}c@{}}Inner Loop\\ Iterations\end{tabular}} & \textbf{\begin{tabular}[c]{@{}c@{}}Optimiser\\ (inner loop)\end{tabular}} & \textbf{$\alpha$} & \textbf{$\beta$} \\ \toprule
Images                 & 128                 & 25                                                                       & SGD                                                                       & 0.001                          & 0.1                           \\
ECG Tasks              & 64                  & 25                                                                       & SGD                                                                       & 0.001                          & 0.1                           \\
Patient Care           & 64                  & 25                                                                       & SGD                                                                       & 0.0001                         & 0.1    \\ \bottomrule                      
\end{tabular}
\end{sc}
\end{small}
\end{table*}

\section{Implementation Details}
\label{sec:app_4}

\subsection*{SCA scores}

 SCA-based affinity matrix $\mathbf{A}$ requires computing Equation 16 of the manuscript that measures the average deviation between task-specific sample-wise optima for a pair of tasks. Each optima (model) is composed $l$ trainable tensors $[\mathbf{M}_1,\mathbf{M}_2,\mathbf{M}_3\ldots\mathbf{M}_l]$. To facilitate the computation of SCA scores, we apply the following operation to transform each optima to a vector: $\text{vec}(\mathbf{M}_1) \oplus \text{vec}(\mathbf{M}_2) \oplus \ldots \oplus \text{vec}(\mathbf{M}_n)$, where $\text{vec}()$ refers to vectorising the tensor and $ \oplus$ refers to concatenation operation.    

In all experiments, we have used this vectorisation step to compute the SCA scores. However, if models are very large, comparing vectorised models/optima might become extremely inefficient. In such scenarios, Equation 16  can be updated to compare corresponding tensors in two task-specific sample-wise optima $\theta^{\tau_i\star}_i$ and $\theta^{\tau_j\star}_i$

\begin{equation}
    a_{\tau_i,\tau_j}= \frac{1}{n} \frac{1}{l} \sum_{i=1}^n \sum_{k=1}^l\lVert \text{vec}\bigl( \prescript{k}{}\theta^{\tau_i\star}_i \bigr) - \text{vec} \bigl( \prescript{k}{}\theta^{\tau_j\star}_i \bigr) \rVert_1,
    \label{eq:pair_tensor}
\end{equation}

where $\prescript{k}{}\theta^{\tau_i\star}_i$ is $k$-th weight tensor of $\theta^{\tau_i\star}_i$. \\

\noindent \textsc{Extension to meta-learning:} As discussed earlier, if we are dealing with one inner iteration of Reptile, the pair-wise SCA affinity between two tasks $\tau_1$ and $\tau_2$ can be computed as:

\begin{equation}
    a_{\tau_1,\tau_2}= \frac{1}{n} \sum_{i=1}^n\lVert \theta^{\tau_1\star}_i - \theta^{\tau_2\star}_i \rVert_1, 
    \label{eq:pair_rep}
\end{equation}

where $\theta^{\tau_1\star}_i$ and $\theta^{\tau_2\star}_i$ are task-specific sample-wise optima obtained from samples $(\mathbf{x}_i^{\tau_1},y_i^{\tau_1})$ and $(\mathbf{x}_i^{\tau_2},y_i^{\tau_2})$, respectively.

\subsection*{GAT architecture and training details} We use a simple architecture consisting of two graph attention layers that intakes node embeddings $\mathbf{F} \in \mathbb{R}^{T \times T}$(SCA-based pairwise affinity with each of $T$ nodes in graph) and tries to reconstruct $\mathbf{F}$. Here $T$ is the number of tasks. The architecture used in this work is as follows:\\

\noindent $\mathcal{G}(V,E,\mathbf{F}) \rightarrow \textrm{Graph Attention($T$ units, $2$ attention heads)}\\\rightarrow\mathbf{Z}\rightarrow \textrm{Graph Attention($T \times 2$ units, $1$ attention head)}\rightarrow \mathbf{F}'$\\

\noindent We use a MSE loss function and Adam with a fixed learning of $0.001$ to train GAT the model for $100$ epochs. From trained model, we obtain $z$-embeddings, $\mathbf{Z} \in \mathbb{R}^{T \times 2T}$ where each row is a representation of corresponding node in input graph.  \\

\subsection*{GMM for Task groupings}

\textsc{\textbf{GMM-based Clustering for obtaining task-groupings:}}
We apply GMM-based clustering on $z$-embedding for soft clustering, with the log-likelihood for node features $\mathbf{Z}$ expressed as:

\begin{equation}
\mathcal{L}(\mathbf{Z}) = \sum_{i=1}^{T} \log \left( \sum_{k=1}^{K} \pi_k \mathcal{N}(z_i | \mu_k, \Sigma_k) \right).
\end{equation}

Here, $T$ is the number of tasks or nodes, $K$ is the number of clusters, $\phi_k$ is the mixing coefficient of $k$-th mixture in GMM, and \( \mathcal{N}(\mu_k, \Sigma_k) \) represents the Gaussian distribution for $k$-th cluster. The soft assignment of $i$th node to $k$th cluster/mixture is calculated by:

\begin{equation}
r_{ik} = \frac{\pi_k \mathcal{N}(\mathbf{z}_i | \mu_k, \Sigma_k)}{\sum_{j=1}^{K} \pi_j \mathcal{N}(\mathbf{z}_i | \mu_j, \Sigma_j)}
\end{equation}

We can use Expectation-Maximisation algorithm to compute parameters, i.e.~$\pi_k$, $\mu_k$ and $\Sigma_k$, of GMM. \\

As discussed in the main text, we implement an additional refinement step aimed at assigning singular clusters (clusters containing only one node) to the closest or most similar cluster, thereby forming a new cluster that replaces the singular one.\\

\noindent \textsc{\textbf{Quality control:}} To assess quality of clustering, we calculate the silhouette score for each node. The silhouette score is a measure of how similar a node is to other nodes in its own cluster compared to nodes in other clusters. It provides a clear indication of the cohesion and separation of the clusters. The calculation of this score involves determining the average intra-cluster distance $a(i)$ and the smallest average inter-cluster distance $b(i)$ for each node:

\begin{equation}
s(i) = \frac{b(i) - a(i)}{\max\{a(i), b(i)\}}
\end{equation}

Here, \( s(i) \) represents the silhouette score for node \( i \), with a value ranging from -1 to +1. A high silhouette score indicates that the node is well matched to its own cluster and distinct from neighbouring clusters.

We then compute the average silhouette score across all nodes, which serves as an indicator of the overall quality of the clustering:

\begin{equation}
\bar{s} = \frac{1}{T} \sum_{i=1}^{T} s(i).
\end{equation}

In this work, we perform run task grouping module (GAT training and clustering) $10$ times and select the clustering configuration that provides maximum average silhouette score. 

\section{Training setups}
\label{sec:app_5}

We use the same setup to train models for task groupings identified by the proposed framework as well as the baseline methods.

\subsection*{MTL Scenario}
As discussed in the main text, we evaluate all task grouping approaches using hard parameter sharing neural architectures that are characterised by a set of shared layers followed by task-specific prediction layers). For QM9 dataset, graph convolutional neural network used in Nash MTL has also been used here. This architecture can be found at \url{https://github.com/AvivNavon/nash-mtl/blob/main/experiments/quantum_chemistry/models.py}.

Similarly, for Celeb-A dataset, we have used model used in TAG. This model can be found at \url{https://github.com/google-research/google-research/blob/master/tag/celeba/CelebA.ipynb}.

Parameter setting used for both these datasets is documented in Table \ref{tab:mtl_pars}.

To implement GradNorm, we used alpha=0.1 and 0.05 for Celeb-A and QM9 dataset, respectively. These values were tuned over the search space $\{0.01,0.05,0.1, 0.5, 1.0, 1.5, 2.0, 3.0, 5.0\}$ to provide best performance on validation sets.   

For Nash MTL, we have used a scaled up version where we update task weights after $100$ iterations.

\begin{algorithm}[t]
\caption{Reptile-based framework for training shared as well as task-specific parameters.}
\label{algo:1}
\begin{algorithmic}

\STATE {\bfseries Input:} $\mathcal{D}_t$: Dataset for task $t$, $\theta$: shared parameters, $\phi_t$: task-specific parameters for task $t$, $\alpha$: outer learning rate, $\beta$: inner learning rate
\vspace{0.1cm}

\FOR{$t=1$ {\bfseries to} $T$}
    \STATE $\mathbf{W}_t=\theta$
    \STATE $\mathcal{B}\leftarrow$\textsc{Sample-Batches}($\mathcal{D}_t$)
    \FOR{all $(\mathbf{b},\mathbf{l}) \in \mathcal{B}$}
        \STATE $\ell=\mathcal{L}(f_{\mathbf{W}_t,\phi_t}(\mathbf{b}),\mathbf{l})$
        \STATE $\mathbf{W}_t = \mathbf{W}_t - \beta \nabla_{\mathbf{W}_t} \ell$
        \STATE $\phi_t = \phi_t - \beta \nabla_{\phi_t} \ell$
    \ENDFOR
\ENDFOR
\STATE $\mathbf{G} = \frac{1}{T} \sum_{t=1}^{T} (\theta - \mathbf{W}_t)$
\STATE $\theta = \theta - \alpha \mathbf{G}$

\end{algorithmic}
\end{algorithm}

\subsection*{Gradient-based Meta-learning Scenario}
Similar to MTL setups, we use architectures characterised by shared and task-specific layers. We use Reptile  as outlined by to perform shared optimisation of shared layers, whereas task-specific layers are trained normally. Algorithm \ref{algo:1} documents the process of employing Reptile for this setup.

Table \ref{tab:rep} documents the parameters used for image category, ECG prediction tasks and patient care tasks (additional experiments).

\begin{figure*}[t]
    \centering
    \subfloat[t-SNE representations of sample-wise converged models\centering]{{\includegraphics[scale=0.25,trim={0.5cm 0.5cm 0 0}]{figs/images_cifar_svhn_rec.pdf}}}%
    \qquad
    \subfloat[SCA scores\centering]{{\includegraphics[scale=0.75,trim={0.5cm 0 0 0.5cm}]{figs/sca.pdf} }}%
            \qquad
    \subfloat[TAG\centering]{{\includegraphics[scale=0.75,trim={0.5cm 0 0 0.5cm}]{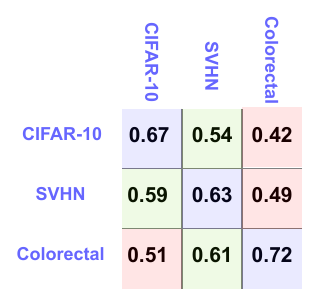} }}%
    
    \caption{\textbf{(a)} t-SNE representations of sample-wise converged \emph{Resnet-50} models for \emph{CIFAR-10}, \emph{street view house number (SVHN)} and \emph{colorectal histology} datasets. \textbf{(b)} The proposed \emph{SCA} and \textbf{(c)} \emph{task-affinity groupings (TAG)}-based task similarities computed among these datasets. Both \emph{SCA} and \emph{TAG} based scores signify that \emph{CIFAR-10} and \emph{SVHN} are more similar to each other than \emph{colorectal histology} dataset.}%
    \label{fig:1}%
\end{figure*}

\begin{figure*}[h]
    \centering
    \subfloat[Dipole moment\centering]{{\includegraphics[scale=0.25,trim={0.5cm 0.5cm 0 0.5cm}]{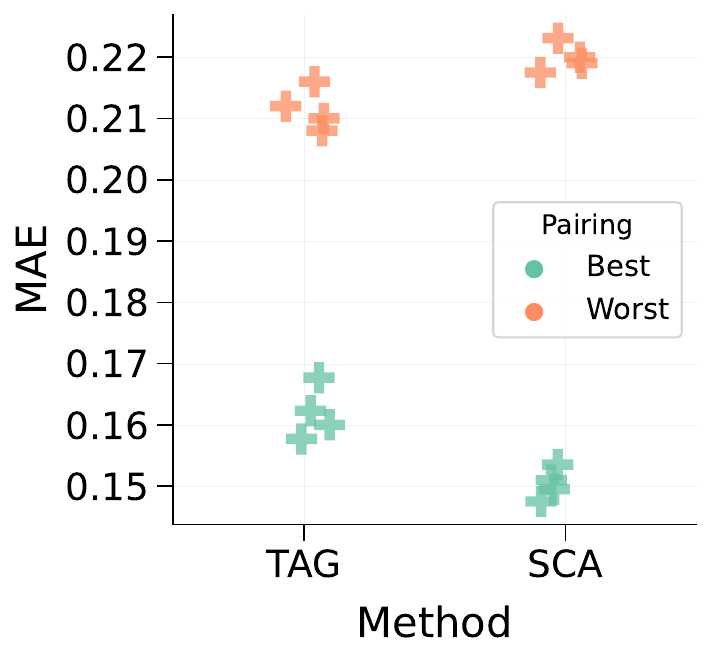}}}%
    \quad
    \subfloat[Isotropic polarizability\centering]{{\includegraphics[scale=0.26,trim={0.5cm 0.5cm 0 0.5cm}]{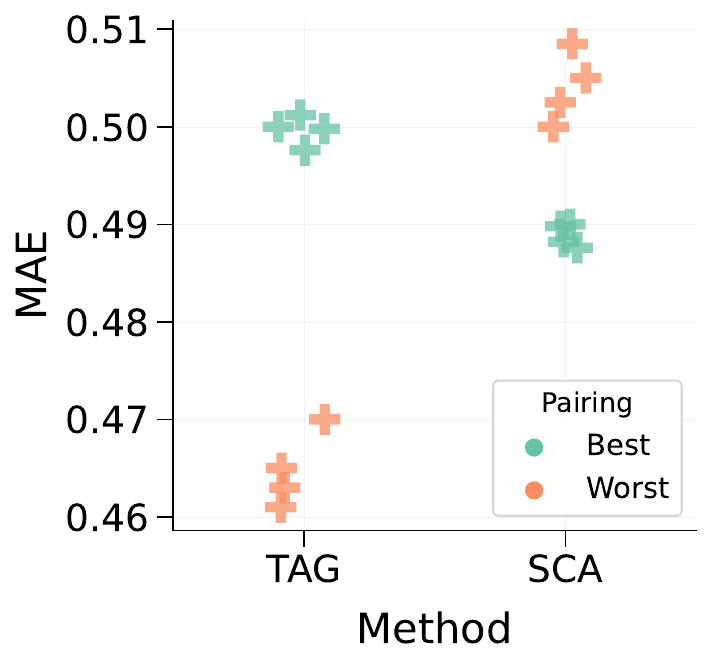} }}%
            \quad
    \subfloat[Atomization free energy at 298.15K\centering]{{\includegraphics[scale=0.26,trim={0.5cm 0.5cm 0 0.5cm}]{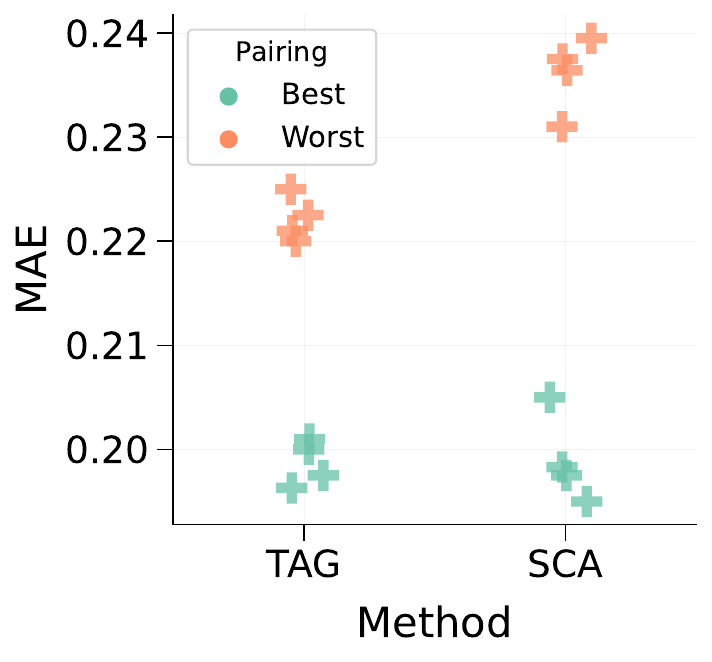} }}%
                \quad
    \subfloat[Electronic spatial extent\centering]{{\includegraphics[scale=0.26,trim={0.5cm 0.5cm 0 0.5cm}]{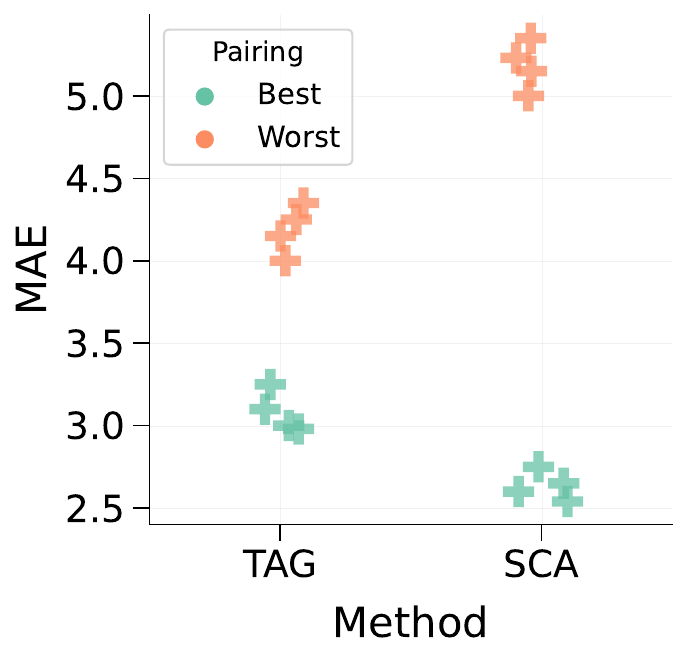} }}%
                \quad
    \subfloat[Zero point spatial extent\centering]{{\includegraphics[scale=0.26,trim={0.5cm 0.5cm 0 0.5cm}]{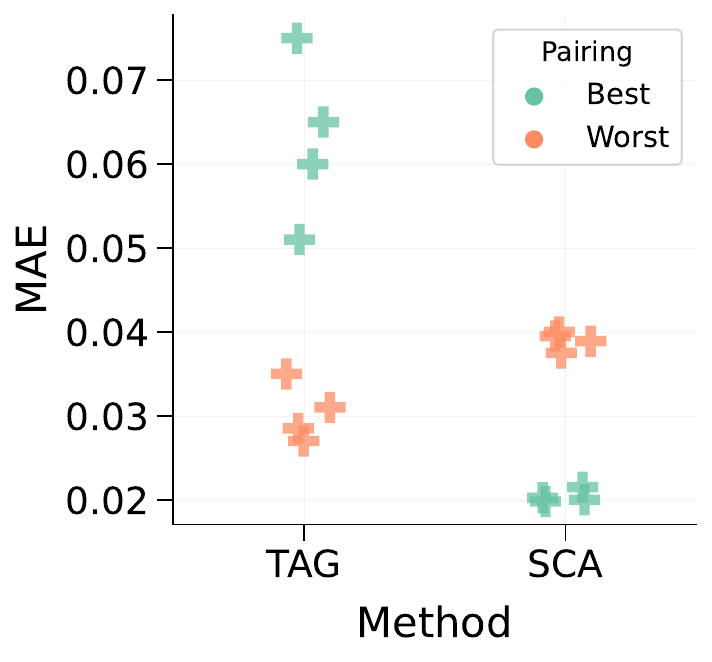} }}%
    
    \caption{Performance of different QM9 tasks in a multi-task (two task learning) setup when paired with the closest (best) and the farthest (worst) candidate task selected using task affinity scores (\emph{TAG}) and the proposed \emph{SCA}-based task affinity metric.}%
    \label{fig:qm9_tag_sca}%
\end{figure*}

\section{Consistency between TAG and SCA scores}
In many cases, we observed a clear consistency between SCA scores and pairwise affinity computed as per TAG. Fig.~\ref{fig:1} illustrates  t-SNE representation of $50$ task-specific sample-wise optima for CIFAR, SVHN and colorectal histology dataset. This figure also depicts the pairwise task affinity computed using TAG and the proposed method. By analysing t-SNE plot, it is clear that CIFAR-10 is closer to SVHN that colorectal histology. This behaviour has been captured by both TAG and SCA.

To further analyse the consistency between TAG and SCA, we selected 5 tasks randomly from QM9 dataset. For each task, we selected most similar and different task from the remaining tasks using SCA and TAG. Then, we trained models on best (similar) and worst (dissimilar) task pairs.
The results of this experiment are depicted in Fig~\ref{fig:qm9_tag_sca}.